\pdfoutput=1
\documentclass[11pt]{article}

\usepackage[table,xcdraw]{xcolor}

\usepackage{tikz}
\usepackage{collcell}
\usepackage{etoolbox}

\newcommand*{\MinNumber}{0.0}%
\newcommand*{\MidNumber}{100.0} %
\newcommand*{\MaxNumber}{10000.0}%

\newcommand{\gradientBLIMP}[1]{
    \ifdim #1 pt > \MidNumber pt
        \pgfmathparse{max(min(
        100.0*
        min(
        0.8*(#1 - \MidNumber)/((\MaxNumber-\MidNumber)/2), 
        0.5*(#1 - \MidNumber)/(\MaxNumber-\MidNumber)+0.5
        )
        ,100.0),0.00)} %
        \hspace{-0.33em}
        \xdef\tempa{\pgfmathresult}
        \cellcolor{black!35!green!\tempa!white!}{#1}
    \else
        \pgfmathparse{max(min(
        100.0*
        min(
        0.8*((\MaxNumber-\MidNumber) - #1)/((\MaxNumber-\MidNumber)/2),
        0.5*(\MidNumber - #1)/(\MidNumber-\MinNumber)+0.5
        )
        ,100.0),0.00)} %
        \hspace{-0.33em}
        \xdef\tempa{\pgfmathresult}
        \cellcolor{red!\tempa!white}{#1}
    \fi
 }

\definecolor{low}{rgb}{0.8, 1, 1}
\definecolor{medium}{rgb}{0.5, 0.7, 1}
\definecolor{high}{rgb}{0, 0, 1}

\usepackage[final]{acl}
\usepackage{tcolorbox}
\usepackage{booktabs, multirow} % for borders and merged ranges
\usepackage{soul}% for underlines
\usepackage{xcolor,colortbl} % for cell colors
\usepackage{changepage,threeparttable} % for wide tables
\usepackage{svg}
\usepackage{csquotes}
\usepackage{times}
\usepackage{latexsym}
\usepackage{graphicx}
\usepackage{tikz}
\usepackage{booktabs}
\usepackage{bm}

\usepackage{tablefootnote}
\usepackage{tabularx}
\usepackage{hyperref}
\usepackage{url}
\usepackage{amsmath}
\usepackage[T1]{fontenc}
\usepackage[utf8]{inputenc}

\usepackage{microtype}

\usepackage{inconsolata}

\usepackage{graphicx}

\title{LongTail-Swap: benchmarking language models' abilities on rare words.}

\author{Robin Algayres* and Charles-Éric Saint-James* and Mahi Luthra* and \\ \textbf{Jiayi Shen* and Dongyan Lin* and Youssef Benchekroun* and Rashel Moritz*} \\ \textbf{and Juan Pino* and Emmanuel Dupoux*+} \\ 
  * Meta AI, + EHESS \\
  \texttt{robinalgayres@meta.com,dpx@meta.com}\\}

\begin{document}
\maketitle
\begin{abstract}
%Typical LM benchmarks measure language performance in the head of the distribution (frequent words, frequent linguistic constuctions, etc). Little is known about the model performance in the long tail of the 

%Despite approaching human-level intelligence, current large language models\textendash trained on internet-scale datasets\textendash still exhibit critical blind spots compared to humans. To address this issue, the BabyLM Challenge is a research initiative focused on training language models on human-scale corpora. In order to contribute to this research line, 

Children learn to speak with a low amount of data and can be taught new words on a few-shot basis, making them particularly data-efficient learners. The BabyLM challenge aims at exploring language model (LM) training in the low-data regime but uses metrics that concentrate on the head of the word distribution. Here, we introduce LongTail-Swap (LT-Swap), a benchmark that focuses on the tail of the distribution, i.e., measures the ability of LMs to learn new words with very little exposure, like infants do. LT-Swap is a pretraining corpus-specific test set of acceptable versus unacceptable sentence pairs that isolate semantic and syntactic usage of rare words. Models are evaluated in a zero-shot fashion by computing the average log probabilities over the two members of each pair.
We built two such test sets associated with the 10M words and 100M words BabyLM training sets, respectively, and evaluated 16 models from the BabyLM leaderboard. Our results not only highlight the poor performance of language models on rare words but also reveal that performance differences across LM architectures are much more pronounced in the long tail than in the head. This offers new insights into which architectures are better at handling rare word generalization. We’ve also made the code publicly available on GitHub, enabling the generation of LT-Swap benchmarks based on any English text corpus.\footnote{\label{sharednote}https://github.com/facebookresearch/lt-swap}.
\end{abstract}

\section{Introduction}

%% DPX: i'd start with a quick intro on the power laws and zipf laws in language
The most recent efforts in NLP have focused on large language models (LLMs) trained on Internet-scale datasets \cite{openai2024gpt4,grattafiori2024llama3,geminiteam2024gemini}. These datasets are composed of trillions of words, which is orders of magnitude above what humans hear and read in their lifetime \cite{willits2021words,hart2003words} yet LLM intelligence still has significant blind spots compared to humans \cite{benchekroun2023worldsense}. As an answer to this problem, a recent research effort, called the BabyLM challenge \cite{warstadt2023callbabylm,choshen2024callbabylm2,charpentier2025callbabylm3}, has stimulated interest in the building of data-efficient language models (LM) pre-trained in human-scale datasets. 

In this work, we brought our attention to the ability of LMs to learn words in a few shots, like infants when they learn to speak \cite{carey1978newwords,markson1997newwords}. This is a notoriously difficult task because of the skewed distribution of words which features a \textit{ long-tail} of very rare word types, a phenomenon known as Zipf's law \cite{zipf}. Those long-tail words are ubiquitous in language \cite{popescu2008wordcounts,fan2010hapax} and are known to pose problems for language models training at all scales. For example, \citet{kandpal2023llmlongtail} shows a large performance decline when LMs are asked questions about rare named entities. \citet{shumailov2024collapse} has unveiled a mode collapse when LMs are recursively trained on their own output. More specific to our problem, \citet{dohmatob2024taletails} has pointed out that LM generations trim the long-tail distribution of words by over-representing frequent words. However, even though long-tail words are notoriously hard to learn, the benchmarks used in the BabyLM challenge (BLiMP \cite{warstadt2019blimp}, EwoK \cite{ivanova2024ewok} and SuperGLUE \cite{wang2019superglue}) focus on the other end of the distribution, avoiding most of the long-tail words (see Appendix \ref{appendix:blimp} for more details). Here, we wish to correct this blind spot and build a metric that specifically targets long-tail words. \\
Addressing the difficulty of learning words from the long-tail, we introduce a framework to automatically create a dataset-dependent evaluation task that we called LongTail-Swap (or LT-Swap). LT-Swap measures the syntactic and semantic abilities of pretrained LMs when probed on long-tail words extracted from their pretraining dataset. Our framework is applicable to any text corpus used for LM pretraining and is scalable to LLM corpus. Inspired by the BLiMP \cite{warstadt2019blimp} benchmark, LMs' performance on LT-Swap is measured by their ability to discriminate between two sentences, one of which is making an incorrect use of a long-tail word. We leverage the instruction-following capabilities of recent LLMs \cite{brown2020gpt3} to generate sentence pairs containing the words of interest. These words range from frequent words, seen more than 512 times, down to words seen only once or even harder: never seen inflections of words present in the pretraining set. LT-Swap can be divided into three subtasks: WordSwap, a semantic metric, InflectionSwap, a POS tagging metric, and AgreementSwap , a syntactic metric.

Using the BabyLM datasets, containing, respectively, 10M and 100M words, we automatically create two LT-Swap tasks: LT-Swap10M and LT-Swap100M and use those two tasks to evaluate 16 different LMs pretrained on the BabyLM datasets. The main takeaways of our analysis are the following: 

$\bullet$ First, while the LT-Swap scores align with those of BLiMP, one of BabyLM's benchmarks, they also reveal a frequency effect, specifically, a sharp decline as the model moves from frequent to long-tail words. These trends are expected and serve as a validity check for LT-Swap. \\
\indent $\bullet$ Second, LT-Swap scores show much greater variation across LM architectures in the long-tail compared to the head of the word distribution. This finding can help identify more data-efficient LM architectures.\\
\indent $\bullet$ Third, we realized that increasing the number of words during pretraining (from 10M to 100M words) without increasing model size makes LMs better at understanding long-tail words \footnote{Note that this is not trivial as it means that on average, words with frequency count $c$ in BabyLM10M are not as well encoded by LMs than words with the same frequency count $c$ in BabyLM100M.}.\\
\indent $\bullet$ Finally, we show that a simple RAG-like method \cite{lewis2021rag}, without any finetuning, can boost semantic scores (WordSwap) for almost all models. However, this method does not improve syntactic performance as evaluated by InflectionSwap and AgreementSwap. This shows that LMs trained on small corpora have in-context-learning abilities, and points to directions in which models could be improved to address the long-tail problem.

These findings highlight the significance of evaluating language models on long-tail words. While the trends observed with LT-Swap are anticipated, our benchmark is the first to enable the measurement of long-tail generalization, addressing a key blind spot in current LM benchmarks. Additionally, we’ve made the code publicly available on GitHub, allowing users to generate new LT-Swap benchmarks based on any English text corpus\footnote{While Wordswap can run on any language, some parts of the code for AgreementSwap and InflectionSwap are hard-coded for the English language only.}. In addition, we release the generated sentences from LT-Swap10M and LT-Swap100M and encourage people to use those tasks to evaluate any LMs pretrained on BabyLM text datasets.
Note that the proposed framework is applied to the text dataset composing the BabyLM challenge but that it should be applicable to any text corpus used for LM pretraining.

\begin{figure*}[h!]
        \centering
        \includegraphics[width=1\textwidth]{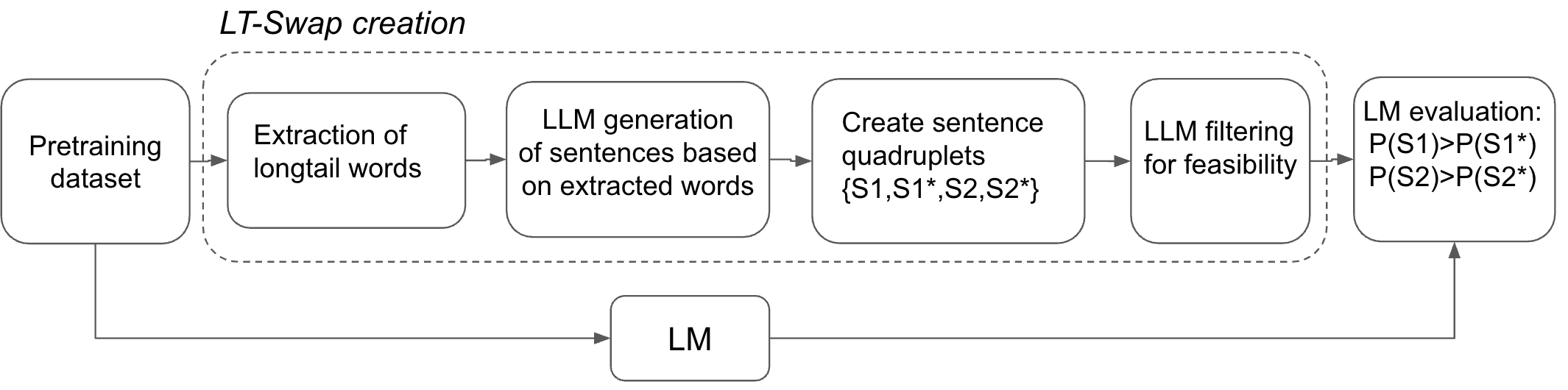}
        
        \caption{Overview of our framework to create the dataset-dependant LT-Swap tasks. The candidate LM is evaluated on sentences, generated by LLM, that contain long-tail words from its pretraining  set. S1* and S2* are obtained by swapping the long-tail words from S1 and S2, thereby creating incorrect sentences. A last LLM filtering step assert that S1* and S2* are indeed less likely than S1 and S2.}
\label{fig:overview}
\end{figure*}

\section{Method}\label{sec:method}

\subsection{Overview}

Figure \ref{fig:overview} gives an overview of our process to create the LT-Swap benchmark automatically starting from a pretraining text dataset. LT-Swap 

First, words are POS tagged with NLTK \cite{birdloper2004nltk} and only verbs and nouns are kept because other syntactic categories are not present in enough instances in the long-tail. We add white spaces between words and symbols\footnote{mainly numbers and punctuations} and segment sentences using white spaces. Word frequencies are obtained by counting the number of occurrences after segmentation. For each word, we automatically compute their most common inflections leveraging the predictable syntactic structure of the English language: plural/singular forms for nouns, past/present-continuous/present for verbs (see Appendix Section \ref{appendix:preprocessing} for details). Selected words and inflections are placed in their respective frequency bins defined by $\{[0],[2^0,2^1[,[2^1,2^2[...[2^n,+\infty[\}$ with $n=9$. The frequency bin $0$ serves for inflections never seen in the pretraining set. From typical work on word counts from psycholinguistics \cite{baayen1995lowfrequency,fan2010hapax,popescu2008wordcounts}, \textit{ long tail word} is a loose term that refers to words below 10ppm\footnote{parts per million} that represent an absolute frequency count up to 100 for a corpus of 10M words. Our choice of frequency bins enables us to explore both the long tail while keeping some frequency bins for the frequent words. See Appendix Table \ref{table:word_examples} for example words that belong to each bin.

After this preprocessing step, we ask an LLM to generate sentences that are example usages for each long-tail word or inflection. The generated sentences are paired together. For each pair of sentences, for example, S1 and S2, we create S1* and S2* by swapping long-tail words. S1 and S2 are meant to be two correct usages of two long-tail words, whereas S1* and S2* are two incorrect sentences. Inspired by \cite{warstadt2019blimp,nguyen2021zrs}, pretrained LMs are evaluated using their own confidence to discriminate the correct from the incorrect sentences. We make sure that the task is feasible with a last LLM filtering step that rejects pairs of sentences that cannot be discriminated from their incorrect versions. The LLM used in this work for both the generation and filtering steps is \textit{Llama3.1-405B}. The way sentences are generated, paired together, and filtered depends on the LT-Swap subtasks: WordSwap, InflectionSwap, AgreementSwap. The rest of this section explains how those three subtasks are created (see example generations in the Appendix Table \ref{table:sentence_examples}). \\

The final LT-Swap score is the accuracy score on the number of times the candidate LM correctly discriminated a correct sentence from an incorrect one. More specifically, we compute an accuracy for each frequency bin and each subtask (WordSwap, InflectionSwap and AgreementSwap) and then average all those accuracies together to form the final LT-Swap score.
\subsection{WordSwap}\label{sec:wordswap}

WordSwap is a task that aims to measure the semantic \textit{} abilities of LMs across the frequency bins. First, we simply ask the LLM to generate sentences using the words selected previously (not the automatic inflections). Some generated sentences may contain words never seen in the pretraining dataset, which can mislead the pretrained LMs when evaluated on our task. Therefore, we filter out generations that include words not present in the pretraining dataset. Then, we pair sentences together if the two long-tail words from which they originate satisfy three conditions: 1) they belong to the same frequency bins, 2) they have the same POS tag, and 3) their frequency bin is not 0. From each pair of generated sentences, we create two semantically incorrect sentences by swapping the target words at the same position but on the other sentence of the pair. The role of the LM is to use its own likelihood to discriminate semantically correct sentences from semantically incorrect ones. Below is a toy example of a quadruplet, S1 and S2 being the generated sentences and S1* and S2* being the incorrect sentences (see Appendix \ref{table:sentence_examples} for other examples).

S1: The \textit{cat} is sleeping on the mat. \\
\indent S2: The \textit{boat} is sailing on the sea \\
\indent S1*: The \textit{boat} is sleeping on the mat. \\
\indent S2*: The \textit{cat} is sailing on the sea\\

At this stage, we need to ensure that the task is feasible. In fact, for words seen very few times, there may not be enough context in the pretraining data set to understand the meaning of those words. In such cases, it would be impossible for any LM trained on this dataset to discriminate correct from incorrect sentences. Therefore, we use the in-context learning abilities of the LLM to play a game in which three sentences are given: a pair of sentences generated previously and one sentence extracted from the pretraining data set that contains one of the two long-tail words. Here is an example: let us use our example generations from before, S1 and S2, and let us pretend that the pretraining dataset contains the sentence 'The cat slept peacefully on the windowsill'. The prompt for the example would be the following:

\begin{tcolorbox}[colframe=blue!25, colback=blue!5, coltitle=black, title=Prompt for WordSwap filtering step]
I have invented a new English word "blick" that you can use as in the following sentence: \\
"<start of sentence> The blick slept peacefully on the windowsill <end of sentence>" \\
Now, I give you two new sentences A and B:\\
"<start of sentence A> The blick is sleeping on the mat. <end of sentence A>"\\
"<start of sentence B> The blick is sailing on the sea <end of sentence B>"\\
Which of the sentences A or B uses the word 'blick' correctly? Put your answer, A or B, in between brackets.
\end{tcolorbox}

In this example, if the LLM is successful, then it means that the sentence extracted from the pretraining set brings enough contextual information on the word 'blick' to tell which of sentence A or B is more probable. To reduce false positives, we duplicate this prompt by switching the order of sentences A and B. Then we also do the same thing using a sentence that contains the word 'boat' from the pretraining set. Finally, the quadruplet, in this case \{S1,S2,S1*,S2*\}, is kept only if the LLM successfully solves the \textit{four} prompts; otherwise, the pair of sentences is discarded.

\subsection{InflectionSwap}\label{sec:inflswap}

InflectionSwap works similarly to WordSwap but aims at measuring the \textit{ syntactic} abilities of pretrained LM instead of \textit{ semantics}. In the previous sections, we asked the LLM to generate sentences using the selected long-tail words, leaving the inflection aside. We keep those sentences for InflectionSwap and, in addition, the LLM is asked to generate sentences using the inflections. Then, we pair sentences together if the two words from which they originate satisfy one condition: they are inflections of each other. From each pair of sentences, we create two syntactically incorrect sentences by swapping the target words at the same position but on the other sentence of the pair. LMs are evaluated in their ability to discriminate syntactically correct and incorrect sentences. Below is an example with S1 and S2 generated by the LLM using \textit{sleep} and \textit{sleeping}.

S1:  He couldn’t \textit{sleep} last night. \\
\indent S2: The baby was \textit{sleeping} peacefully. \\
\indent S1*: He couldn’t \textit{sleeping} last night. \\
\indent S2*: The baby was \textit{sleep} peacefully.\\

Previously, each pair of words was composed of two words belonging to the same frequency bin. Here, it is not possible to do that anymore as there are not enough pairs of words and inflections that both belong to the same frequency bin. Therefore, pairs are put into the frequency bin given by the least frequent of the two words. In this case, the frequency bin can be 0 as during the automatic generation of inflections, we may end up with word types that never occur in the pretraining corpus.

Finally, as for WordSwap, we need to make sure that the InflectionSwap task is feasible by filtering out the sentence pairs that cannot be discriminated by the LLM. Yet, InflectionSwap being simpler than WordSwap, we assume that even if the long-tail words have been seen only once in the pretraining dataset, it is enough context to understand their syntactic function. Therefore, we use the following prompt on every generated sentence pair, for instance here we S1 and S1*. 

\begin{tcolorbox}[colframe=blue!25, colback=blue!5, coltitle=black, title=Prompt for InflectionSwap filtering step]
Given the two sentences A and B:\\
"<start of sentence A> He couldn’t sleep last night. <end of sentence A>"\\
"<start of sentence B> He couldn’t sleeping last night. <end of sentence B>"\\
Which of the two sentences A or B is syntactically correct? Put your answer, A or B, in between brackets.
\end{tcolorbox}

As before, we duplicate this prompt by switching the order of sentences A and B in order to reduce false negatives. Then we do the same with S2 and S2*. The quadruplet \{S1,S2,S1*,S2*\} is kept only if the LLM is correct on four prompts.

\subsection{AgreementSwap}\label{sec:agswap}

Although InflectionSwap measures if pretrained LMs can identify POS tags, it is not an explicit evaluation of identifiable \textit{syntactic rules}. Here, in the spirit of the BLiMP benchmark, AgreementSwap focuses on analyzing specific syntactic rules that can be easily generated by an LLM: \textit{subject-verb agreements}, \textit{anaphore agreements} and \textit{determinant-noun agreements}. Those three rules require an agreement between the subject of the sentence, that is, either singular, third-person, or plural, and another word in the sentence that can be either a verb, a reflexive pronoun, or a determinant.

Once again, the generation process starts by asking an LLM to generate sentences based on pairs of inflected words. However, here, our aim was to control the difficulty of the task by manipulating the number of words between the two elements involved in the agreement relation. Therefore, we designed both short-distance and long-distance agreement tasks by explicitly instructing the LLM to generate sentences where the subject appeared either immediately adjacent to the agreement target (short-distance) or separated by intervening material (long-distance). After the generation process, quadruplets are made in the same fashion as for WordSwap and InflectionSwap. Here is an example quadruplet for each of the three syntactic rules for the short-distance and long-distance cases.

$\bullet$ Subject-verb agreement:\\
\indent S1: The \textit{strategist} analyzes.\\
\indent S2: The \textit{strategists} analyze.\\
\indent S1*: The \textit{strategists} analyzes.\\
\indent S2*: The \textit{strategist} analyze.\\
\indent $\bullet$ Anaphore agreement:\\
\indent S1: The \textit{interviewees} considered themselves.\\
\indent S2: The \textit{interviewee} presented herself.\\
\indent S1*: The \textit{interviewee} considered themselves.\\
\indent S2*: The \textit{interviewees} presented herself.\\
\indent $\bullet$ Determinant-noun agreement:\\
\indent S1: This \textit{renunciation}.\\
\indent S2: These \textit{renunciations}.\\
\indent S1*: This \textit{renunciations}.\\
\indent S2*: These \textit{renunciation}.\\
\indent $\bullet$ Subject-verb agreement long-distance:\\
\indent S1: The \textit{strategist} that can be trusted, analyzes.\\
\indent S2: The \textit{strategists} that can be trusted, analyze.\\
\indent S1*: The \textit{strategists} that can be trusted, \\ \indent analyzes.\\
\indent S2*: The \textit{strategist} that can be trusted, analyze.\\ \\
\indent $\bullet$ Anaphore agreement long-distance:\\
\indent S1: The \textit{interviewees} that can be chosen for \\ \indent the job, 
 considered themselves.\\
\indent S2: The \textit{interviewee} that can be  trusted, \\ \indent presented herself.\\
\indent S1*: The \textit{interviewee} that can be chosen for \\ \indent the job, considered themselves.\\
\indent S2*: The \textit{interviewees} that can be trusted, \\ \indent presented herself.\\

Note that we did not do a long-distance version of the determinant-noun agreement because it was too difficult to control this setup with the LLM (see more examples in the Appendix Table \ref{table:sentence_examples}). The prompts used to generate AgreementSwap sentences are in Appendix \ref{sec:appendix_agprompts}. \\
Once all sentences are generated, we need to make sure that swapping the target word will indeed be a test of the target syntactic rule. For instance in the determinant-noun task, the LLM may generate S1:'This \textit{misconduct} is a serious offense' and S2:'These \textit{misconducts} are serious offenses'. The last words 'offense'/'offenses' give away the singular/plural nature of the subject. By swapping the long-tail words in italic font, this would not be a pure determinant-noun agreement task but also a subject-verb agreement task. Therefore, for all agreement tasks, we usually cut the sentence just after the last word marking the agreement. In this case the pair S1:'This \textit{misconduct}.' and S2:'These \textit{misconducts}.'. Even though cutting the sentences makes them not semantically correct, the candidate LM should still realize that one sentence is less likely than the other.
%\footnote{Most pairs in the BLiMP benchmark also do not sound semantically correct yet are used to evaluate syntactic performances of pretrained LMs.}. 
Finally, we make sure that all quadruplets can be solved by an LLM using the same filtering strategy as for InflectionSwap. In addition, we check programatically that generated sentences correctly follow the desired syntactic rules.
\begin{figure*}[h!]
        \centering
        \includegraphics[width=.5\textwidth]{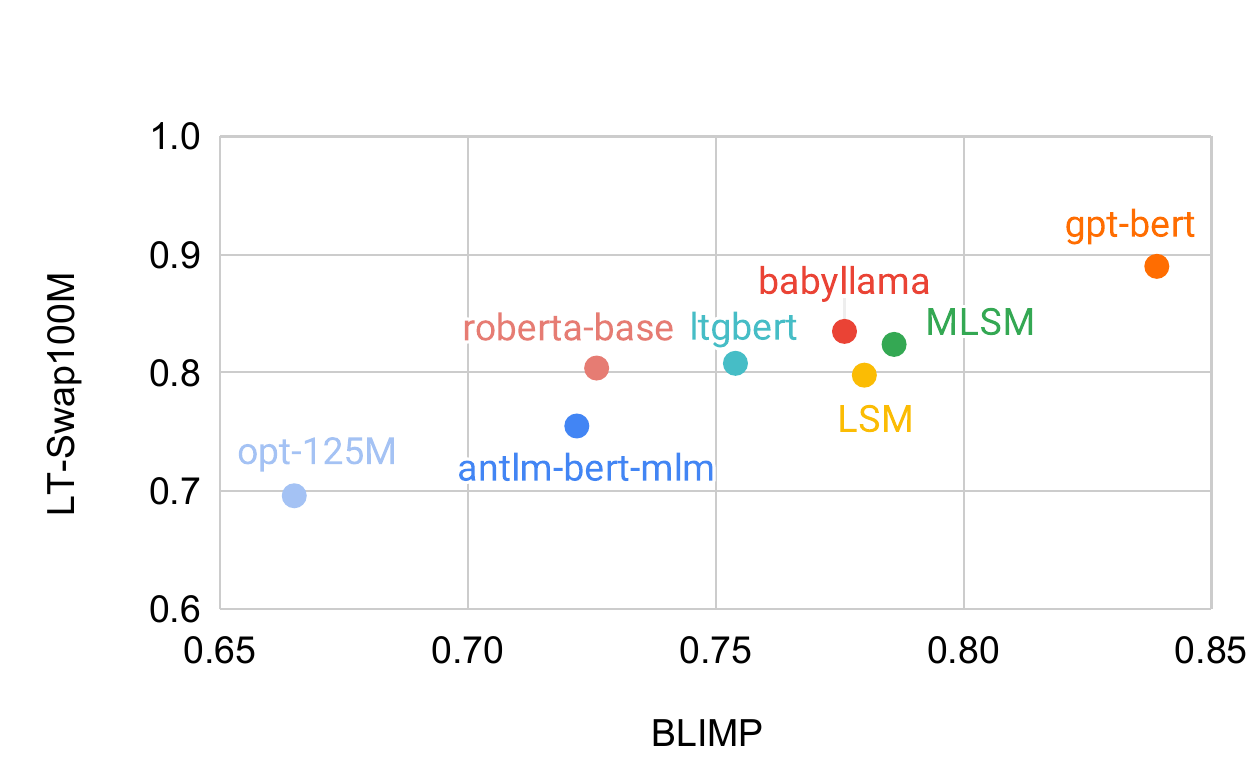}\hfill
        \includegraphics[width=.5\textwidth]{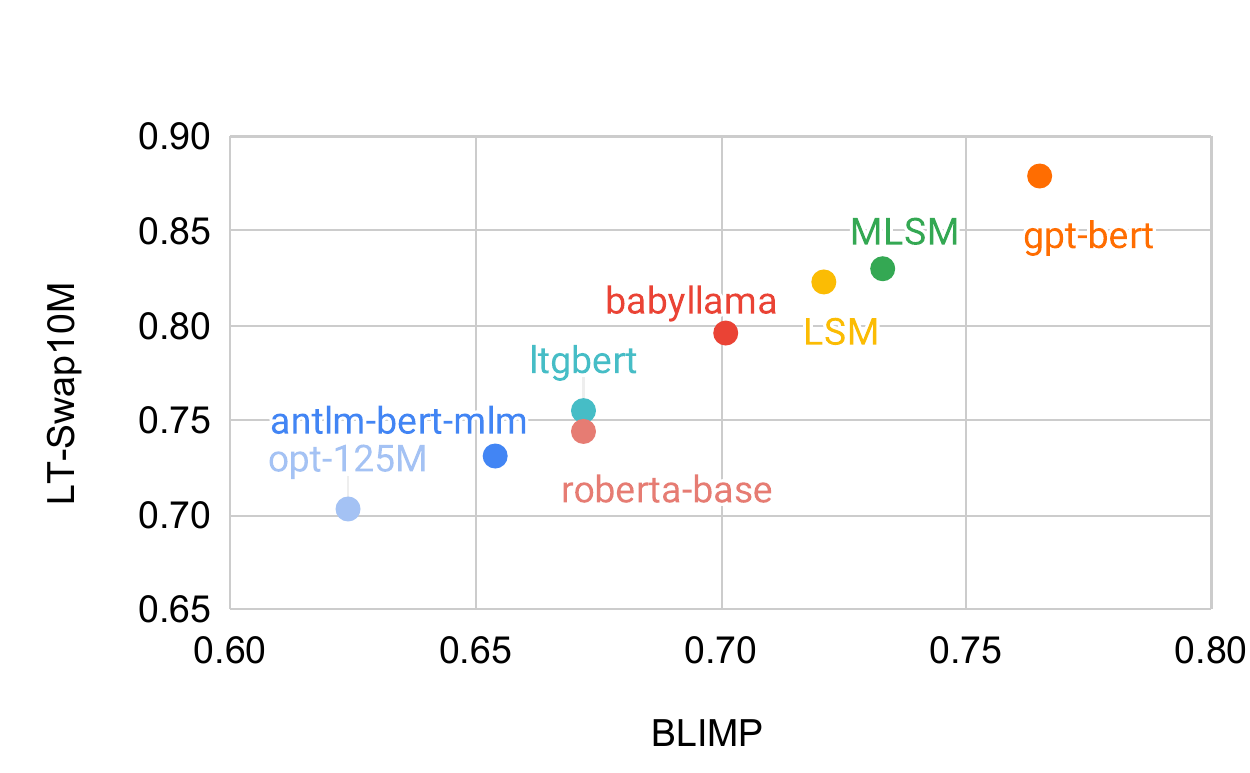}
        
        \caption{BLiMP and LT-Swap scores across LMs pretrained on BabyLM10M or BabyLM100M. Both LT-Swap and BLiMP are accuracy scores with 50\% random chance.}
\label{fig:swap_vs_blimp}
\end{figure*}

\begin{figure*}[h!]
        \centering
        \includegraphics[width=.42\textwidth]{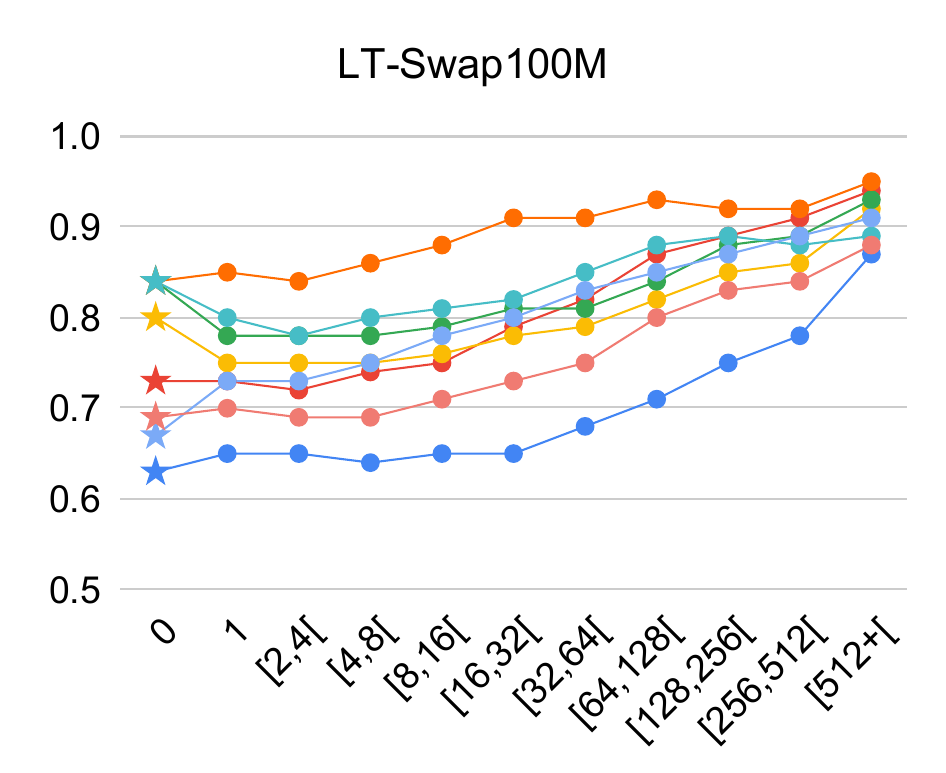}\hfill
        \includegraphics[width=.55\textwidth]{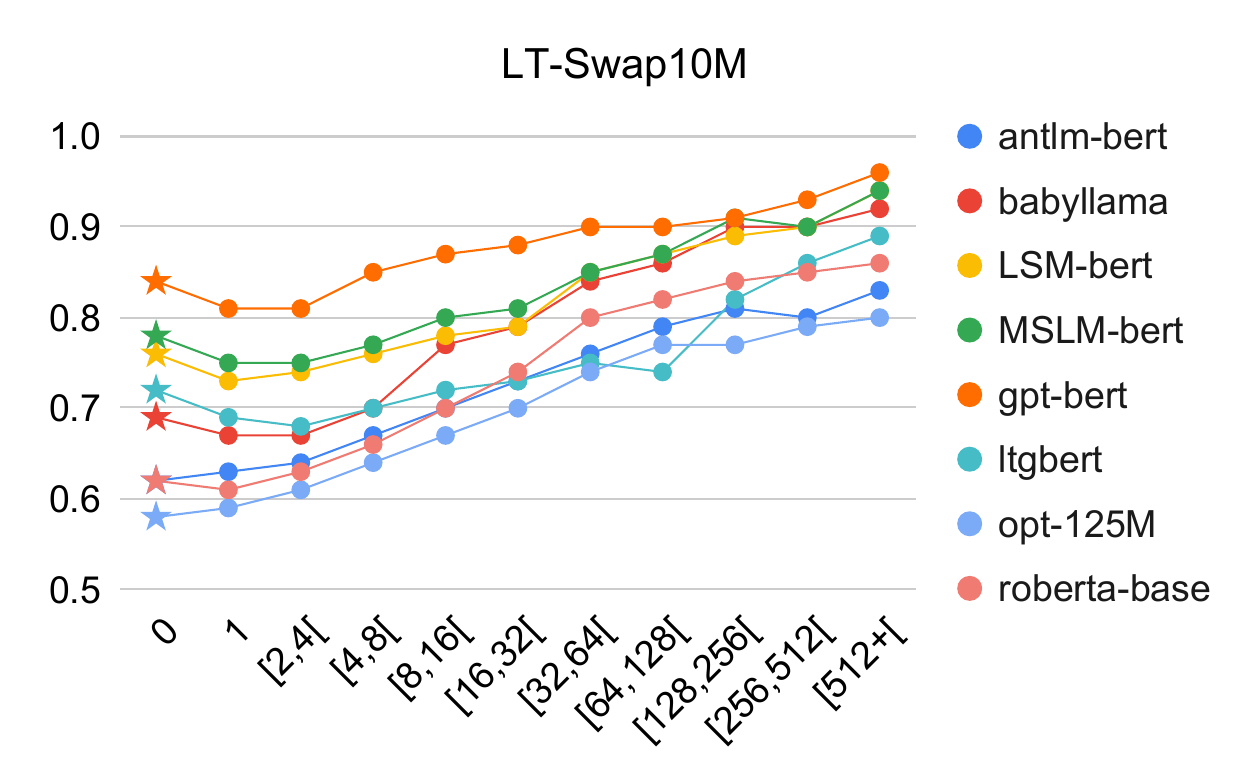}
        \caption{LT-Swap scores (i.e average of WordSwap, InflSwap and AgrSwap) broken down across frequency bins for selected architetures pretrained on BabyLM10M and BabyLM100M (16 models in total). The star symbols is a reminder that the bin 0 is only the average of AgreementSwap and InflectionSwap which explains why it is shifted up compare to the bin 1. The standard error for the Swap scores is less than 0.02 while for BLiMP the standard error than go as high as 0.2.}
\label{fig:swapscores_per_freqs}

\end{figure*}

\section{Experiments}

\subsection{Swap10M and Swap100M}

We applied our task creation framework from Section \ref{sec:method} to the two BabyLM text corpora composed of 10M words, called BabyLM10M, and 100M words \cite{warstadt2023callbabylm}, called BabyLM100M. Those corpora are a mixture of written text (English Wikipedia, children stories \cite{gerlach2018gutenberg}), transcribed dialogues (Switchboard\cite{stolcke2000switchboard}, Opensubtitles\cite{lison2016opensubtitles}) \\ and transcribed child-directed speech \cite{maxwhinney2000childes}. We refer to Swap10M and Swap100M for the tasks built on the word distribution of BabyLM10M and BabyLM100M respectively (see the number of pairs per frequency bins in Appendix Tables \ref{tab:nb_pairs_swap10M} and \ref{tab:nb_pairs_swap100M}). 

%The values of the frequency bins are set to be in the powers of two from 1 to 512 with additional bin 0 for AgreementSwap and InflectionSwap. POS tags are obtained using the NLTK toolkit. Sentences are generated and filtered using the largest LLM we had at disposal, Llama3.1-405B, to ensure the best generations and the highest filtering quality. For WordSwap, we make sure the LLM to generate sentences of at least 20 words, which is long enough so that there is enough context to understand the target word. For the last filtering step in WordSwap, the success rate of the game is around 40\% for all frequency bins.

\subsection{Selected range of pretrained LMs}

We evaluated all language model architectures submitted to the BabyLM Challenge, each available in two versions: one pretrained on the BabyLM10M dataset and the other on the BabyLM100M dataset. We have two GPT models (OPT \cite{zhang2022opt}, BabyLLama\cite{warstadt2023babylmfindings}), one BERT model (Roberta-base\cite{liu2019roberta}), one modified BERT with disentangled attention and shared positional embeddings (LGT-BERT\cite{samuel2023ltg}), two mixtures of GPT and BERT (GPT-BERT\cite{charpentier2024gptbert}, ant-lm-mlm\cite{yu2024antlm}), and two BERTs trained with a teacher-student method (LSM and MLSM\cite{berend2023mlsm}). All models use either BPE or a SentencePiece tokenizer; they have around 150 million parameters. Each model is evaluated using the appropriate inference method following the recommendations of their respective published papers (see the Appendix \ref{appendix:confidence_score} for details on inference methods and models not included in this study).

\begin{table}[!htp]\centering
\resizebox{\linewidth}{!}{
\scriptsize
\begin{tabular}{lcccc}\toprule
  &LT-Swap &WS &IS & AS \\\cmidrule{2-5} 
10M &1 (0) &1 (0) &0.98 (0) &0.90 (0.04) \\\cmidrule{1-5}
100M  &1 (0) &1 (0) &0.93 (0) &0.48 (0.35) \\
\bottomrule
\end{tabular}
}
\caption{Spearman correlation cofficients and p-value (in between parenthesis), for both dataset sizes 10M and 100M, when correlating frequency bins and Swap scores. Those coefficients are computed on each model separately, and then averaged across models. WS: WordSwap, IS: InflectionSwap, AS: AgreementSwap.}\label{tab:correlations}
\vspace{-1.em}
\end{table}

\begin{table*}[!htp]\centering
\resizebox{0.7\textwidth}{!}{
\scriptsize
\begin{tabular}{lcccc|cccc}\toprule
&\multicolumn{4}{c}{Accuracy drop} &\multicolumn{4}{c}{Accuracy spread ratio} \\\cmidrule{2-9}
&LT-Swap &WS &IS &AS &LT-Swap &WS &IS &AS \\\cmidrule{2-9}
10M &-0.208 &-0.319 &-0.199 &-0.126 &1.29 &1.44 &1.80 &0.90 \\\cmidrule{1-9}
100M &-0.163 &-0.291 &-0.175 &-0.079 &2.11 &3.80 &2.38 &1.26
 \\
\bottomrule
\end{tabular}
}
\caption{Accuracy drop and Accuracy spread ratio of LT-Swap scores (i.e average of WordSwap, InflSwap and AgrSwap) for 10M and 100M dataset sizes. Both metric are differences/ratios between the most frequent and the least frequent bins. WS: WordSwap, IS: InflectionSwap, AS: AgreementSwap.}\label{tab:perfdrop_spreadratio}

\end{table*}

\vspace{-0.5em}
\section{Results}
\vspace{-0.5em}
In this section, we present the scores obtained by the selected range of language models on our tasks, along with a detailed analysis of the results.

\subsection{LT-Swap frequency effect}\label{sec:lt-swap-correlations}

We show in Figure \ref{fig:swap_vs_blimp} that LT-Swap is consistent with a similar benchmark, BLiMP, which also evaluates the confidence of LMs in distinguishing syntactically correct sentences from incorrect ones. In addition to being correlated with BLiMP scores\footnote{We recomputed BLiMP scores ourselves due to some inconsistencies on the BabyLM leaderboard and papers.}, LT-Swap scores shown in Figure \ref{fig:swapscores_per_freqs} reveal a frequency effect with a sharp decline of LMs' performance when going from frequent to rare words (see Appendix Tables \ref{tab:scores_per_pos_10M} and \ref{tab:scores_per_pos_100M} and Figure \ref{fig:scores_per_freqs} detailed scores per subtasks). This frequency effect is made possible because, unlike the BLiMP benchmark, LT-Swap includes enough sentence pairs per frequency bin to ensure that the standard error for all models across all bins remains below 0.02 absolute points (see Appendix \ref{appendix:blimp} for an analysis of BLiMP scores across frequencies).

Although LT-Swap shows a clear correlation with frequencies in Figure \ref{fig:swapscores_per_freqs}, the breakdown of LT-Swap into its subtasks in Table \ref{tab:correlations}, shows that this is not the case for AgreementSwap. These findings suggest that the acquisition of syntactic patterns in language models is not primarily driven by absolute word frequencies, but rather by other factors such as model architecture, tokenization strategy, and loss function. For example, if the tokenization method separates base words from their plural markers (e.g., "cat" and "-s"), the model may detect syntactic inconsistencies without necessarily recognizing the lexical identity of the base word. This reduces the model’s reliance on word frequency and allows it to leverage subword-level or structural cues instead (see Appendix Section \ref{appendix:tokenization} for an analysis on the impact of tokenizations).

\subsection{Accuracy drop and spread ratio}

From the scores obtained on the LT-Swap benchmark, we make two observations that are summarized in Table \ref{tab:perfdrop_spreadratio}.

First, the sharp drop in accuracy, that is, the average drop in accuracy between the highest frequency bin and the lowest frequent bin, is a sign that LMs struggle with long-tail words much more than with frequent words. Even in the case of hapax, the accuracy scores remain significantly above random chance (50\%), which is a surprising result. We propose two possible explanations for this result. First, the language model (LM) might rely on the meaning associated with each BPE unit, enabling it to infer the meaning of hapax words. Alternatively, it may be the case that the LM requires only a single gradient update to distinguish novel words effectively. Further investigation is needed to better understand what aspects of word learning a LM can acquire from a single exposure.

The second observation is that accuracy scores are more different across models for the rare words than they are for the frequent words. In Table \ref{tab:perfdrop_spreadratio} we measure the spread ratio by taking the difference in LT-Swap scores between the best and worst models in the rarest bin and dividing it by the difference between the best and worst models in the most frequent bin. These ratios can reach up to 3.8 for WordSwap, highlighting the importance of evaluating LMs on very rare words, where performance differences are more pronounced.

\subsection{10M and 100M models}

\begin{figure}[h!]
        \centering
        \includegraphics[width=\linewidth]{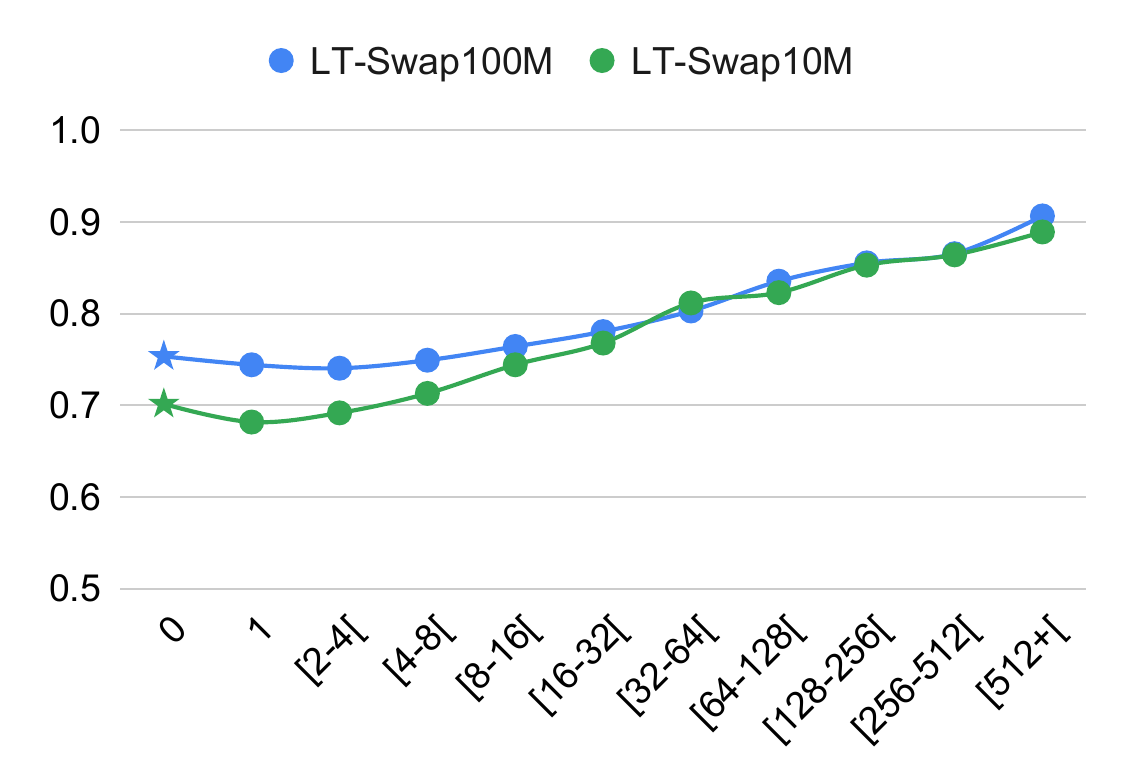}
        \caption{LT-Swap scores averaged across LMs. The standard error on those scores is less than 0.02. The star symbols is a reminder that the bin 0 is the average of only AgreementSwap and InflectionSwap, not WordSwap which does not apply on this bin. }
        \vspace{-1em}
\label{fig:100M_vs_10M}
\end{figure}

In Figure \ref{fig:100M_vs_10M}, we present the average LT-Swap scores for language models (LMs) trained on 10M or 100M words. It is evident that LMs trained on 10M words underperform those trained on 100M words in the lowest frequency bins. This result is noteworthy because it indicates that, on average, words with frequency count $c$ in BabyLM10M are not encoded as well by LMs as words with the same frequency count $c$ in BabyLM100M. This suggests that training on a larger corpus enables LMs to better capture contextual information, thereby improving their ability to understand rare word types.

\begin{figure}[h!]
        \centering
        \includegraphics[width=\linewidth]{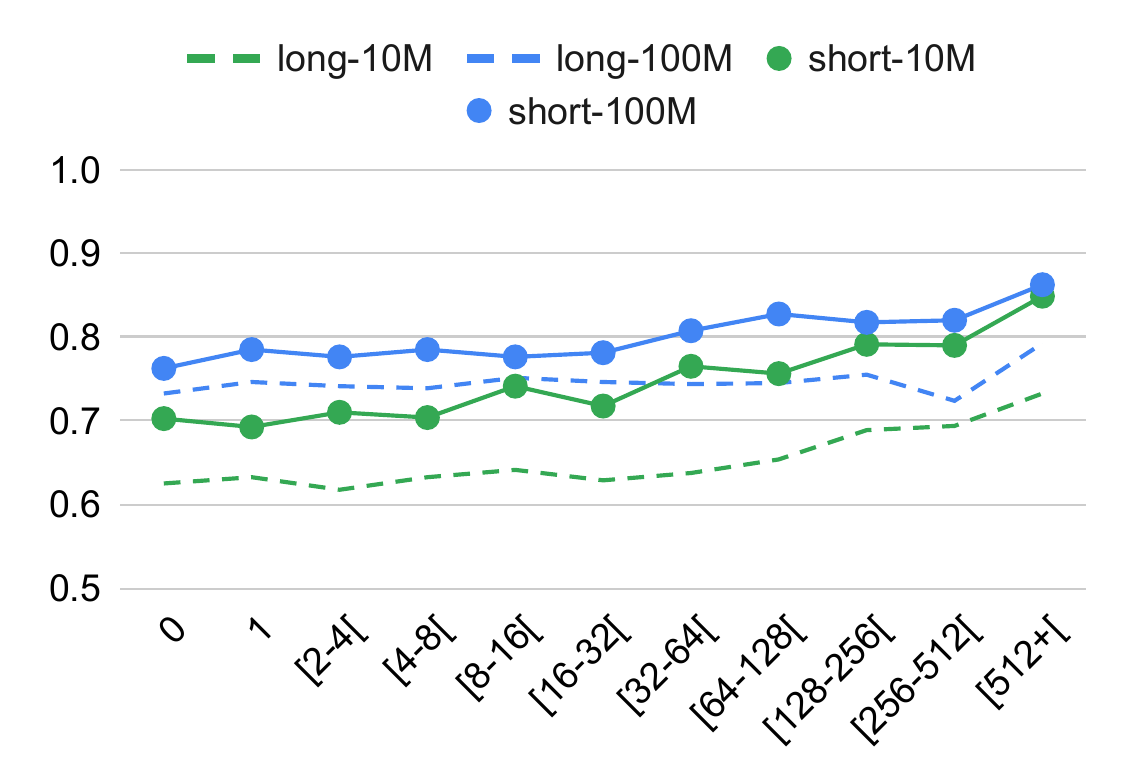}
        \caption{AgreementSwap short and long distance scores averaged over LMs for 10M and 100M pretraining size. The standard error on those scores is less than 0.03.}
        \vspace{-1em}
\label{fig:long_short}

\end{figure}

In addition, in Figure \ref{fig:long_short}, we calculate the long- and short-distance scores of AgreementSwap on LMs trained in 10M or 100M words. As expected, the short-distance case is easier to solve than the long-distance case for both BabyLM datasets. More interestingly, the accuracy gap between short and long distances is smaller for LMs trained on 100M words than for LMs trained on 10M words. This shows that increasing the number of words also helps syntax learning, even for the lowest-frequency bins. \\

Our analysis demonstrates that, while keeping model size fixed\footnote{All models evaluated in this study have approximately 125 million parameters}, increasing the dataset size improves language models (LMs) on the long tail. These findings complement the work of \citet{kandpal2023llmlongtail}, which showed that increasing model size, while keeping dataset size fixed, also benefits performance on the long tail. However, the combined effect of both larger dataset and model sizes remains unclear, and further research is needed to disentangle the individual contributions of dataset size and model size to long-tail performance.

\subsection{Boosting semantic scores using a prefix}

With a simple RAG-like \cite{lewis2021rag} method that requires no finetuning, we show that almost all LMs get a performance boost on the lowest frequency bins of the WordSwap task. The idea is to say that sentences from the pretraining set can be added in the context of WordSwap sentences to add valuable information on the meaning of long-tail words. Therefore, for each WordSwap sentence, both semantically correct and incorrect ones, we retrieve one sentence from the pretraining set that contains the target long-tail word and we add that sentence as a prefix. When computing LMs' confidence score, we exclude the prefix from the (pseudo) log-probabilities. In Table \ref{tab:prefix} we show a consistent positive gain for almost all LMs on the average of the three lowest frequency bins of WordSwap: [1,2[,[2,4[ and [4-8[. This simple experiment suggests that even LMs trained on small datasets have in-context learning abilities.

This method is not intended for general-purpose use, as it typically degrades performance on WordSwap for higher frequency bins and on InflectionSwap and AgreementSwap across all frequency bins. %This makes sense as syntactic tasks will not benefit from semantic are meant to evaluate syntactic abilities which are less sensitive to the addition of more context than a semantic task like WordSwap.

\begin{table}[!htp]\centering

\scriptsize
\begin{tabular}{lcc}\toprule
&WordSwap10M &WordSwap100M \\\cmidrule{2-3}
antlm-bert &+0.10 &+0.13 \\
babyllama &+0.07 &+0.09 \\
LSM-bert &+0.02 &+0.05 \\
MSLM-bert &+0.01 &+0.05 \\
gpt-bert &+0.08 &-0.03 \\
ltgbert &+0.06 &+0.07 \\
opt-125M &+0.09 &+0.06 \\
roberta-base &+0.04 &+0.03 \\ \midrule
\textit{average}  & \textit{+0.049} &\textit{+0.048}\\
\bottomrule
\end{tabular}
\caption{Accuracy increase on WordSwap, averaged over the three lowest frequency bins (1,[2,4[ and [4,8[), due to adding prefix for LMs trained on BabyLM10M and BabyLM100M. For higher frequency bins, this method significantly decrease accuracies. }\label{tab:prefix}

\end{table}

\section{Conclusion}

In this paper, we introduced a framework for generating LT-Swap, a set of tasks based on long-tail words from a pretraining dataset. Our benchmark highlights the need to study the performances of LMs on the long-tail as it unveils sharper differences across architectures. In addition, our benchmark shows that increasing the number of words during improve LMs encoding of long-tail words. Finally, our prefix method hints at a potential direction to improve LMs on long-tail generalizations.

\section*{Limitations}

Our work relies on NLTK POS tagger and on automatically computed word inflections that are hard-coded for the English language. More work is needed to adapt InflectionSwap and AgreementSwap to the multilingual setting. Only WordSwap can be used as is in another language.

Another limitation of our work is that we have focused only on the BabyLM datasets, which contain a lot of transcribed speech. This kind of text has a shorter long tail than written text like Wikipedia. We do not know how the frequency effect would change when using other datasets with significantly different word distributions. In addition, the English language used in this work is mainly US English and is not representative of all English-speaking groups. Finally, one takeaway from the paper is that training on more words increases the ability to understand long-tail words. We do not know how this effect evolves with even larger dataset sizes and how this effect changes for different model sizes.

\section*{Acknowledgments}

This work was granted access to the HPC resources of IDRIS under the allocation 2024-AD011014739R1 made by GENCI, and was supported in part by Agence Nationale de Recherche (ANR-17-EURE-0017 FrontCog, ANR-10-IDEX-0001-02 PSL and ANR-23-IACL-0006 France 2030). 
ED in his EHESS role was funded in par by an ERC grant (InfantSimulator) and Agence Nationale de Recherche (ANR-17-EURE-0017 FrontCog, ANR-10-IDEX-0001-02 PSL). Views and opinions expressed are those of the authors only and do not necessarily reflect those of the European Union or the European Research Council.
Neither the European Union nor the granting authority can be held responsible for them.

% Bibliography entries for the entire Anthology, followed by custom entries
%\bibliography{anthology,custom}
% Custom bibliography entries only

\bibliography{custom}

\appendix

\section{Word candidates}\label{appendix:preprocessing}
% This part is not clear yet... I need to rewrite  it 
The LT-Swap sentences are constructed to be example usages of target words whose frequencies are computed on a given pretraining set. Here are the preprocessing steps to construct the list of word candidates and compute their frequencies.

First, all sentences are segmented into words using white spaces.  All words are POS-tagged in the context of their sentence using NLTK \cite{birdloper2004nltk} before being set to lowercase. Then, we add white spaces on the left and right sides of all symbols and figures. For instance, e.g: \textit{"Jeremy's 59th birthday."} becomes \textit{" jeremy ' s 5 9 th birthday . "}. Finally, we re-segment using the newly added white spaces and remove all words that do not belong to the English dictionary. After those steps, the word frequencies can be computed by counting the number of occurrences. Words are placed in their respective frequency bins which we chose to be the power of twos between 1 and 512. At this stage, we remove all words that are not nouns and verbs, as only those two are present in enough instances in all frequency bins. Specifically, the only POS tags that are kept at this stage are: noun, plural noun, verb, past tense verb, and present continuous verb.\\

We want to account for the fact that some words may belong to the long-tail while having inflections (i.e., a singular, plural, third person, past tense, etc.) that are much more frequent. Such words cannot count as a \textit{real} long-tail words as the concept they refer to may actually be quite frequent. Therefore, for all words selected that way, we automatically compute a set of predefined common inflections. As the English language has a predictable grammar, plural and third-person forms are created by adding 's' at the end of words, present-continuous and past tense by adding 'ing' or 'ed', each time checking if the created inflection belongs to the English dictionary \footnote{For plural and third person, words finishing by 'y' gets '+ies', and words finishing by x,z,s gets +es. For past tense, the words ending with 'y' get '+ied'.}. Based on the frequencies of all inflections of a word, we decide whether this word should be considered for the word pairing stage. More specifically, for a given word, we sum the frequencies of itself and all its inflections. If this sum is larger than the ceiling of this word's frequency bin, the word is filtered out from the candidate list.

\section{LMs selection and confidence scores}\label{appendix:confidence_score}

There are a number of architectures submitted to the BabyLM challenge that we decided not to include in our analysis, either because loading those on huggingface raises an error or because they have not been pretrained on the official versions of BabyLM10M and BabyLM100M. Some models have been pretrained either on only one of the two BabyLM datasets, or on phonemized versions of the BabyLM datasets, or even on totally different datasets (the challenge allows for pretraining on other datasets as long as the maximal number of words does not exceed 100M). \\

The way confidence scores are obtained depends on the LM architecture. GPT-like models are evaluated using the confidence score given by the log-probability of predicting the sentence. It is computed by summing over all tokens in the sentence, the log-softmax for predicting the next token. For BERT models, we compute a pseudo-likelihood score: we mask tokens one at a time and sum the log-softmax output of predicting the masked tokens. More than one token could be masked to obtain a pseudolikelihood score, yet we found that scores were in general lower if more than one token is masked. Finally, for the two special cases of mixture between GPT and BERT, ant-lm-mlm is evaluated like a GPT model, and GPT-BERT is evaluated like a BERT model with the difference that the labels are shifted right in a GPT-like fashion. We did not search the temperature parameters that maximize the scores as suggested in \cite{charpentier2024gptbert}.

\section{Using quadruplets}\label{sec:appendix_quadruplets}

An important aspect of this task is that it is based on quadruplets of sentences instead of simple pairs of sentences as most zeroshot NLP benchmarks (storyclose, blimp, wuggy,...). Using quadruplets instead of pairs limits biases that could nudge the LM in the right direction (sentence lengths, unigram probabilities,...). In Table \ref{tab:quadruplets}, we compute the Spearman correlation of performances with frequency bins, in the same way as in the main paper at Table \ref{tab:correlations}. Yet this time we made pairs instead of quadruplets.\\

\section{AgreementSwap prompts}\label{sec:appendix_agprompts}

Here is the list of the prompts used to generate the AgreementSwap syntactic examples.
342qkj
\begin{tcolorbox}[colframe=blue!25, colback=blue!5, coltitle=black, title=Prompt for Subject-Verb agreement at short distance with 'cat' and 'cats']
Using the nouns 'cat' and 'cats', please write a minimal pair of sentences that show a short distance subject-verb agreement at the present tense. The subject and the verb must be placed close to each other. You must encapsulate the two sentences together in between brackets.
\end{tcolorbox}

\begin{tcolorbox}[colframe=blue!25, colback=blue!5,  coltitle=black, title=Prompt for Subject-Verb agreement at long distance with 'cat' and 'cats']
Using the nouns 'cat' and 'cats', please write a minimal pair of sentences that shows a long distance subject-verb agreement through a relative clause starting by 'that can be'. For instance, using the nouns 'neighbor' and 'neighbors', you can write something like: 'The neighbor that can be trusted lets his dog out. The neighbors that can be trusted let their dog out.'. Now please do the same with 'cat' and 'cats'. You must encapsulate the two sentences together in between brackets.
\end{tcolorbox}

\begin{tcolorbox}[colframe=blue!25, colback=blue!5,  coltitle=black, title=Prompt for Anaphore agreement at short distance with 'cat' and 'cats']
Using the nouns 'cat' and 'cats', please write a minimal pair of sentences that shows a short distance usage of reflexive pronouns. The pronouns must be placed close to the subjects 'cat' and 'cats'. Please use the past tense. Now please do the same with 'cat' and 'cats'. You must encapsulate the two sentences together in between brackets.
\end{tcolorbox}

\begin{tcolorbox}[colframe=blue!25, colback=blue!5,  coltitle=black, title=Prompt for Anaphore agreement at long distance with 'cat' and 'cats']
Using the nouns 'cat' and 'cats', please write a minimal pair of sentences that shows a long distance usage of reflexive pronouns through a relative clause starting by 'that can be'. For instance, using the verbs 'medecine' and 'medecines', you can write something like: 'The medecine that can be bought anywhere, proved itself to be very effective. The medecines that can be bought anywhere, proved themselves to be very effective'. Now please do the same with 'cat' and 'cats'. You must encapsulate the two sentences together in between brackets.
\end{tcolorbox}

\begin{tcolorbox}[colframe=blue!25, colback=blue!5,  coltitle=black, title=Prompt for determinant-noun agreement at short distance with 'cat' and 'cats']
Using the nouns 'cat' and 'cats', please write a minimal pair of sentences that shows a determiner-noun agreement, using either that/these/this/those. For instance, using the nouns 'misconduct' and 'misconducts', you can write something like: 'This misconduct is a serious offense. These misconducts are serious offenses.'. Now please do the same with 'cat' and 'cats'. You must encapsulate the two sentences together in between brackets.
\end{tcolorbox}

\begin{table*}[!htp]\centering

\scriptsize
\begin{tabular}{lrrrrrrrrr}\toprule
&\multicolumn{4}{c}{No quadruplets} &\multicolumn{4}{c}{Using half of the pairs} \\\cmidrule{2-9}
R and p-values &All &WordSwap &InflSwap &Syntax &All &WordSwap &InflSwap &Syntax \\\cmidrule{1-9}
10M &0.86 (0.04) &0.99 (0) &0.56 (0.18) &-0.18 (0.45) &0.99 (0) &1 (0) &0.98 (0) &0.88 (0.02) \\\cmidrule{1-9}
100M &0.7 (0.23) &1 (0) &-0.34 (0.28) &-0.45 (0.4) &0.97 (0) &1 (0) &0.92 (0) &0.51 (0.46) \\\cmidrule{1-9}
\bottomrule

\end{tabular}
\caption{Using quadruplets is necessary to see correlations between LT-Swap performances and with frequency bins}\label{tab:quadruplets}
\end{table*}

\section{BLiMP10M and BLiMP100M}\label{appendix:blimp}

The BLiMP benchmark \cite{warstadt2019blimp} can also be applied to the study of the effect of word frequencies on LMs' syntactic capabilities. As for the LT-Swap tasks, the LMs are evaluated in BLiMP using their own confidence on automatically generated sentence pairs. Most of the time, the pair of sentences differs by only one word, so they can be placed in different frequency bins using the frequencies computed in the BabyLM datasets. As for InflectionSwap and AgreementSwap, the two target words may have different frequencies, in which case the frequency bin for this pair is the one of the least frequent target words. Some BLiMP sentence pairs do not differ by any word but by the order of the words; in these cases, the frequency bin is the least frequent word in the whole sentence. We call BLiMP00M and BLiMP100M the two different sortings of sentence pairs in frequency bins obtained with the frequencies of BabyLM10M and BabyLM100M, respectively. The number of pairs of sentences per frequency bin can be found in Tables \ref{tab:nb_pairs_blimp10M} and \ref{tab:nb_pairs_blimp100M}. 

\begin{figure*}[h!]
        \centering
        \includegraphics[width=.42\textwidth]{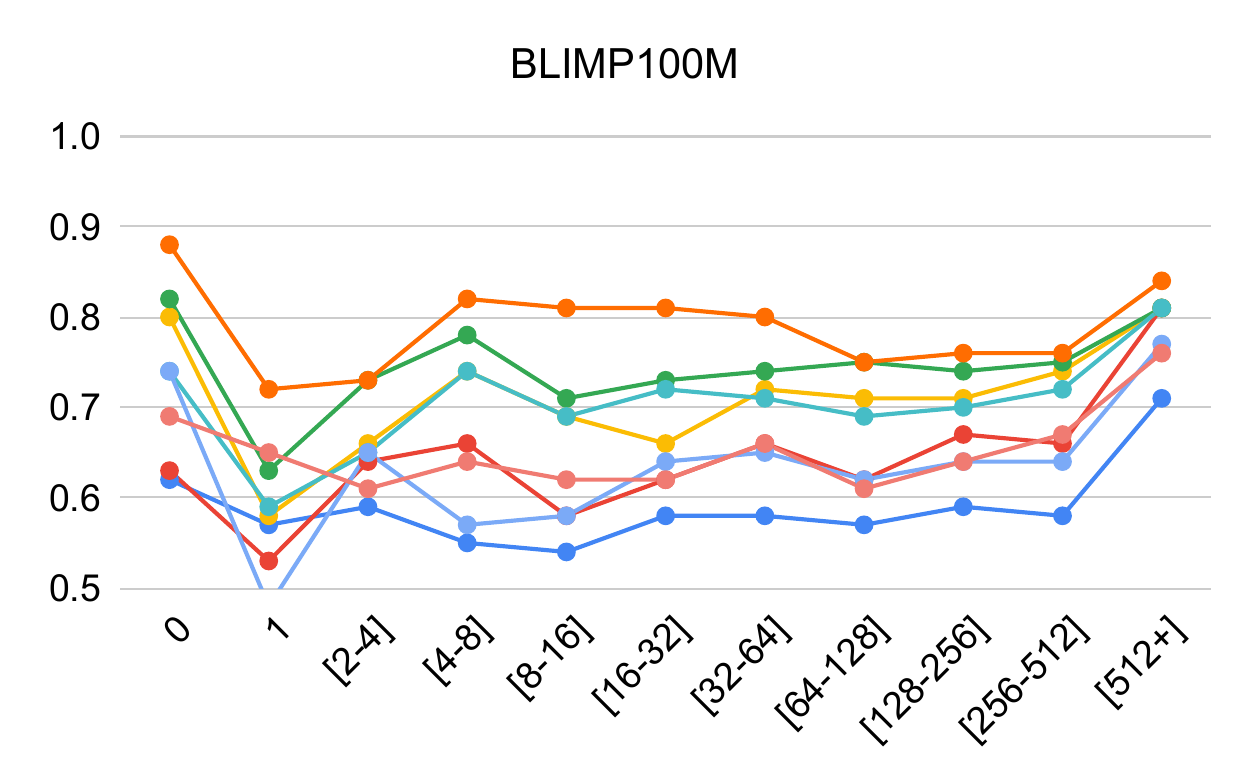}\hfill
        \includegraphics[width=.55\textwidth]{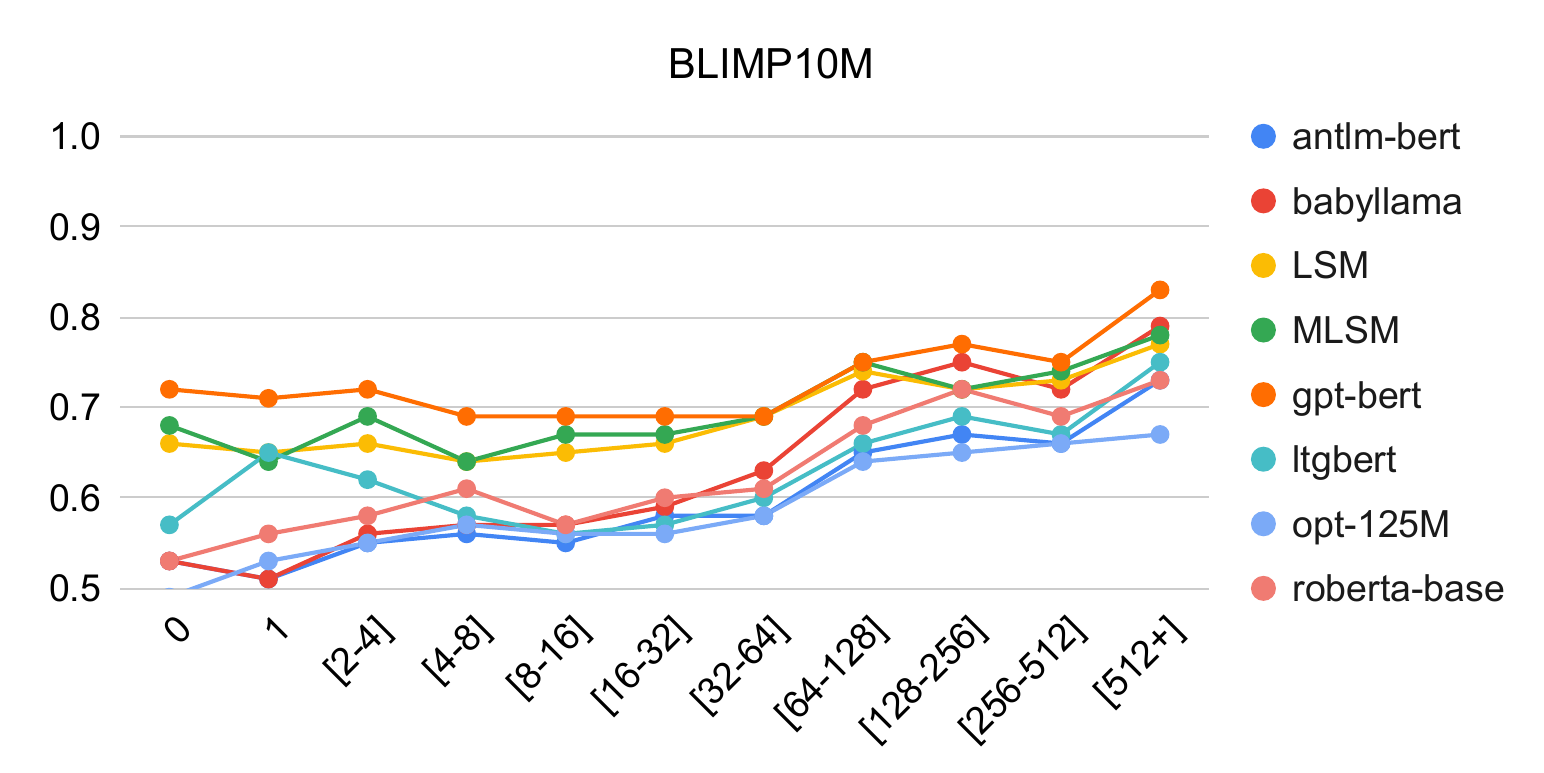}
        \includegraphics[width=.45\textwidth]{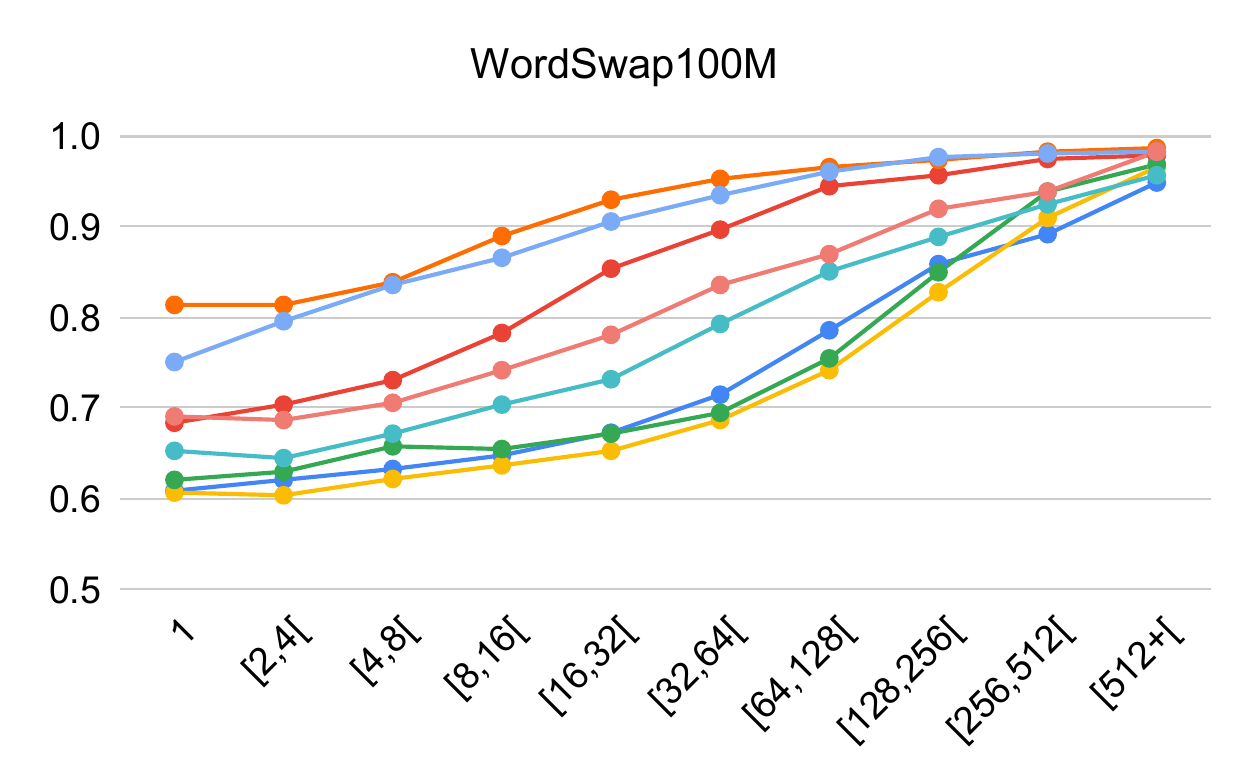}\hfill
        \includegraphics[width=.55\textwidth]{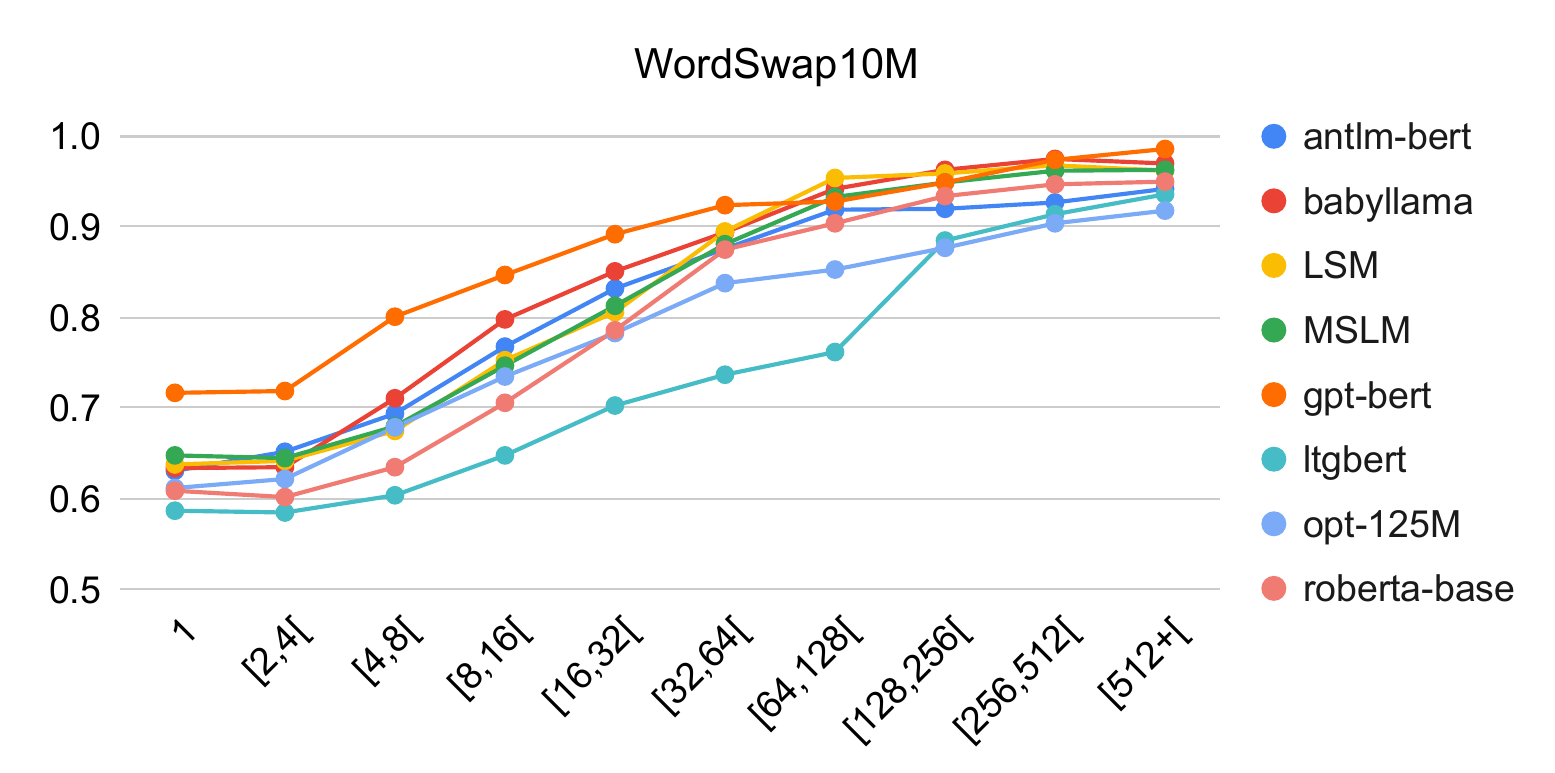}
        \includegraphics[width=.45\textwidth]{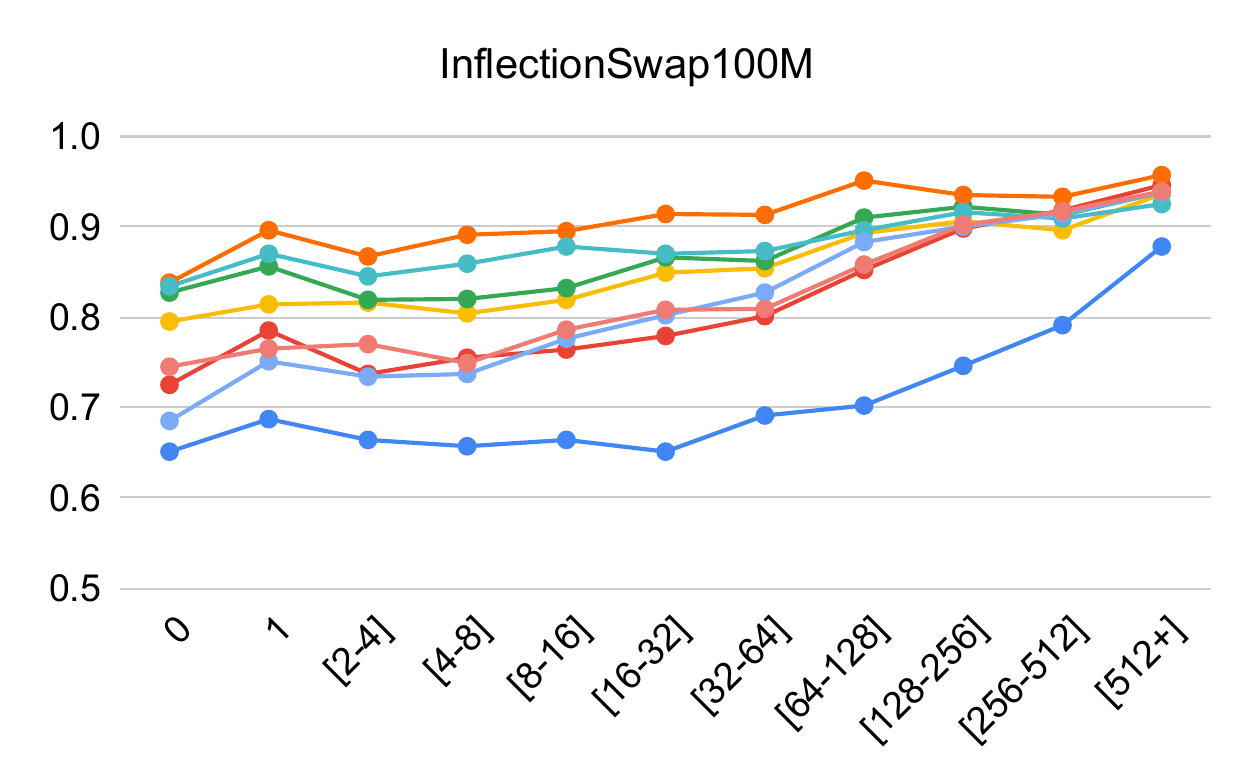}\hfill
        \includegraphics[width=.55\textwidth]{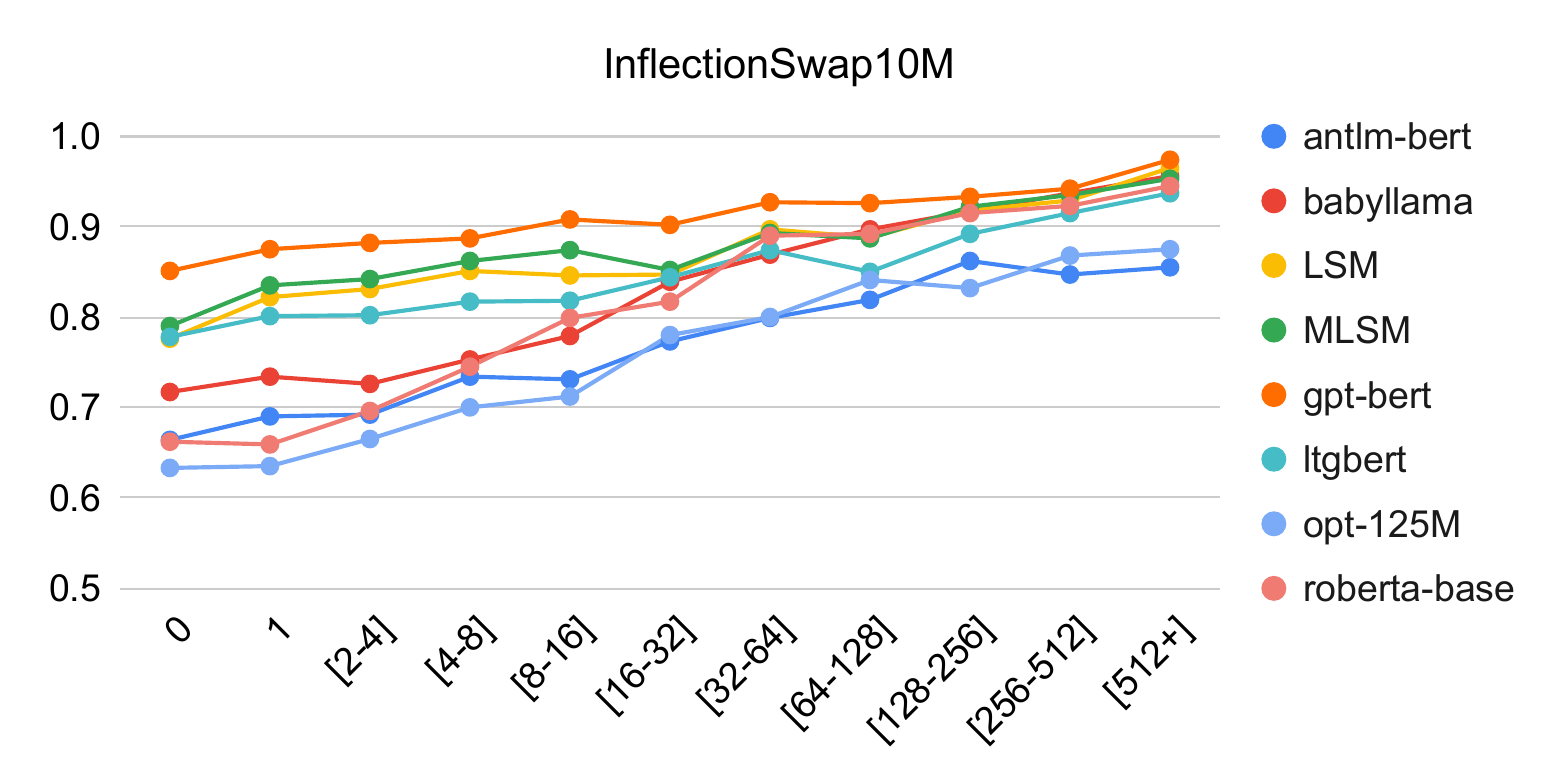}
        \includegraphics[width=.45\textwidth]{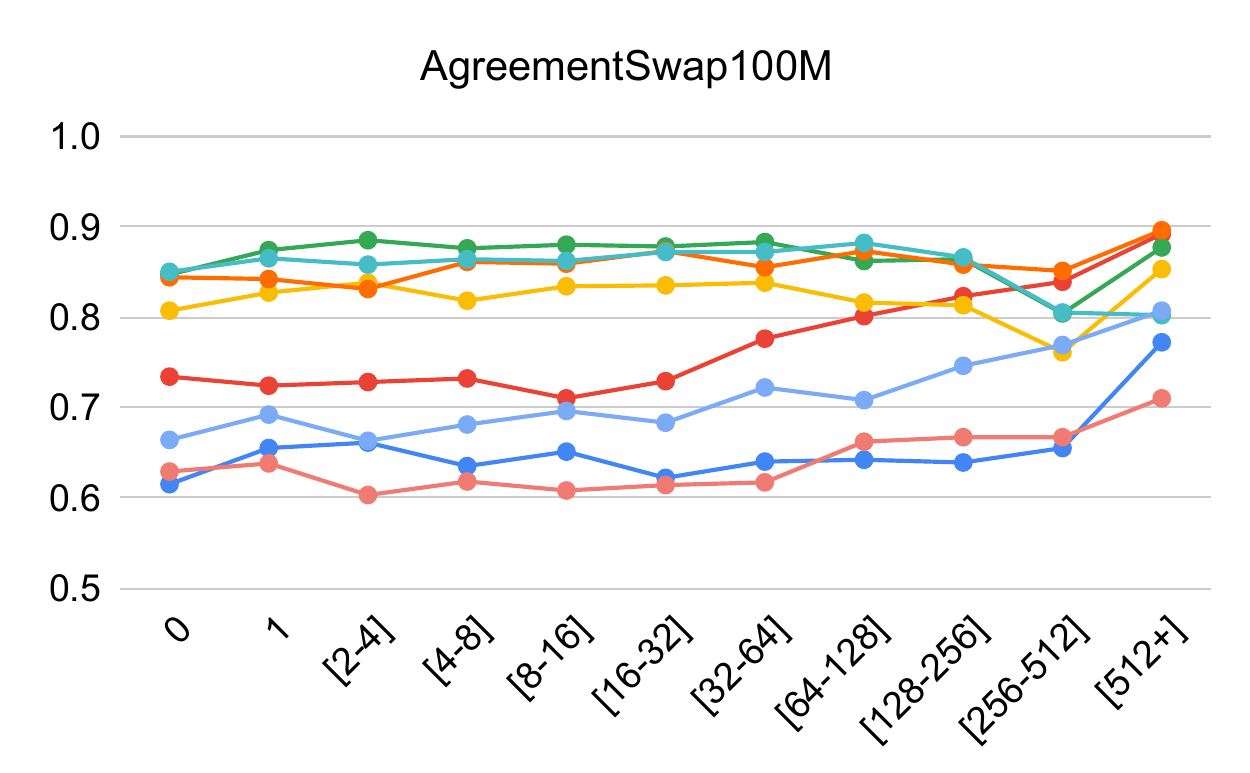}\hfill
        \includegraphics[width=.55\textwidth]{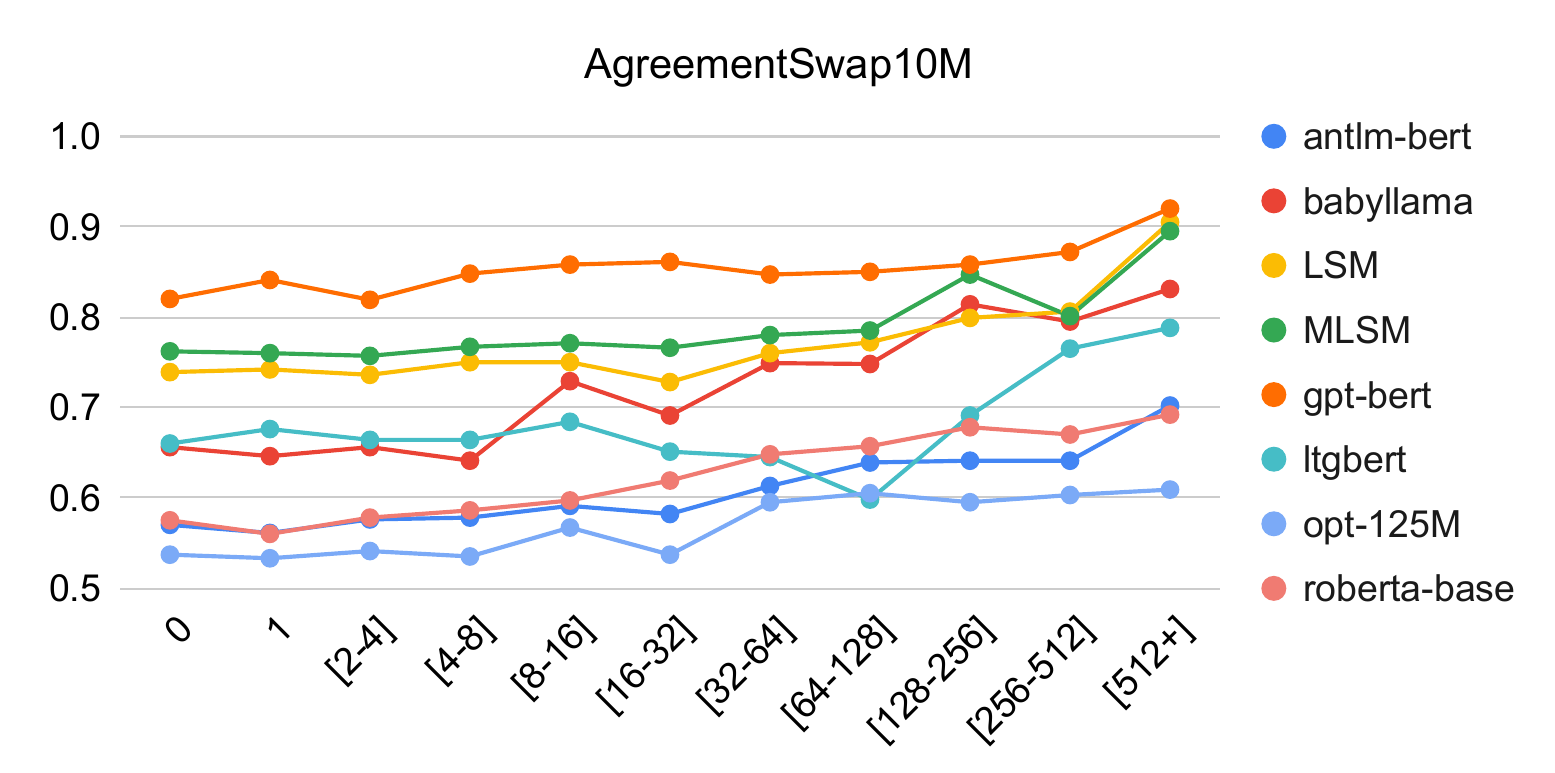}
        \caption{Scores obtained on our three Swap tasks over BabyLM10M and BabyLM100M for the 8 selected architectures across frequency bins. The standard error for all Swap scores is less than 0.02.}
\label{fig:scores_per_freqs}
\end{figure*}

The BLiMP10M and BLiMP100M scores per frequency bin can be found in Figure \ref{fig:scores_per_freqs}. By computing the Spearman correlation with frequencies,  we observe that BLiMP10M is well correlated with frequencies (spearman R=0.89) yet BLiMP100M is not (spearman R=0.29). The low correlation coefficients of BLiMP100M could be a consequence of the extremely imbalanced number of pairs between frequency bins and subtasks, as shown in Tables \ref{tab:nb_pairs_blimp10M} and \ref{tab:nb_pairs_blimp100M}. In contrast, our tasks contain more pairs in subtasks and each frequency bin as shown in Tables \ref{tab:nb_pairs_swap10M} and \ref{tab:nb_pairs_swap100M}. Those results prove that BLiMP is not adapted to the study of long-tail words.

\begin{table*}[!htp]\centering
\resizebox{\textwidth}{!}{
\def\arraystretch{1.2}

\begin{tabular}{crrrrrrrrrrrrrr}\toprule
BLiMP-100M &anap-agr &arg\_struct &binding &control &det-agr &ellipsis &filler &irreg &island &npi &quant &select &sv-agr \\\cmidrule{1-14}
0 &0 &20 &6 &4 &13 &18 &0 &0 &41 &20 &0 &0 &7 \\\cmidrule{1-14}
1 &0 &0 &14 &27 &60 &16 &1 &0 &70 &69 &0 &8 &20 \\\cmidrule{1-14}
2 &0 &7 &0 &32 &28 &14 &1 &0 &23 &30 &0 &9 &13 \\\cmidrule{1-14}
4 &0 &44 &5 &23 &0 &0 &0 &0 &94 &151 &0 &2 &29 \\\cmidrule{1-14}
8 &0 &138 &15 &18 &109 &24 &4 &0 &157 &204 &0 &43 &81 \\\cmidrule{1-14}
16 &0 &281 &35 &94 &361 &69 &5 &0 &320 &403 &0 &43 &305 \\\cmidrule{1-14}
32 &0 &422 &133 &214 &481 &68 &6 &0 &744 &636 &0 &95 &289 \\\cmidrule{1-14}
64 &0 &851 &180 &173 &130 &156 &9 &0 &1265 &975 &0 &138 &154 \\\cmidrule{1-14}
128 &0 &1025 &308 &582 &258 &128 &6 &93 &1270 &994 &0 &145 &278 \\\cmidrule{1-14}
256 &0 &998 &570 &503 &270 &240 &18 &0 &1818 &1349 &0 &292 &341 \\\cmidrule{1-14}
512 &2000 &3212 &5729 &3330 &6290 &1267 &6950 &1907 &2198 &2169 &4000 &1225 &4483 \\\cmidrule{1-14}
\bottomrule
\end{tabular}
}
\caption{Number of pair of sentences in BLiMP across frequency bins using the frequencies of BabyLM100M.}\label{tab:nb_pairs_blimp100M}
\end{table*}

\begin{table*}[!htp]\centering
\resizebox{\textwidth}{!}{
\def\arraystretch{1.2}
\begin{tabular}{crrrrrrrrrrrrrr}\toprule
BLiMP-10M &anap-agr &arg\_struct &binding &control &det-agr &ellipsis &filler &irreg &island &npi &quant &select &sv-agr \\\cmidrule{1-14}
0 &0 &143 &77 &72 &130 &65 &3 &0 &484 &611 &0 &52 &88 \\\cmidrule{1-14}
1 &0 &172 &7 &32 &410 &18 &6 &0 &68 &128 &0 &26 &137 \\\cmidrule{1-14}
2 &0 &326 &75 &166 &281 &67 &5 &0 &542 &534 &0 &83 &244 \\\cmidrule{1-14}
4 &0 &663 &91 &137 &263 &115 &5 &0 &728 &629 &0 &100 &348 \\\cmidrule{1-14}
8 &0 &773 &259 &277 &203 &158 &10 &0 &1201 &956 &0 &123 &183 \\\cmidrule{1-14}
16 &0 &1233 &428 &618 &330 &149 &12 &93 &1714 &1272 &0 &202 &320 \\\cmidrule{1-14}
32 &0 &1027 &643 &679 &320 &247 &22 &0 &1757 &1253 &0 &331 &328 \\\cmidrule{1-14}
64 &0 &977 &437 &745 &469 &286 &24 &0 &897 &790 &491 &277 &355 \\\cmidrule{1-14}
128 &0 &723 &395 &427 &374 &270 &22 &725 &408 &482 &0 &282 &289 \\\cmidrule{1-14}
256 &0 &428 &217 &774 &145 &247 &17 &252 &149 &202 &0 &152 &164 \\\cmidrule{1-14}
512 &2000 &533 &4366 &1073 &5075 &378 &6874 &930 &52 &143 &3509 &372 &3544 \\\cmidrule{1-14}
\bottomrule
\end{tabular}
}
\caption{Number of pair of sentences in BLiMP across frequency bins using the frequencies of BabyLM10M.}\label{tab:nb_pairs_blimp10M}
\end{table*}

\begin{table*}[!htp]\centering
\resizebox{0.8\textwidth}{!}{
\def\arraystretch{1.1}
\begin{tabular}{crrrrrrr}\toprule
LT-Swap-100M  &WS-VERB &WS-NOUN &IS-VERB &IS-NOUN &AG-LONG &AG-SHORT \\\cmidrule{1-7}
0 &NA &NA &1039 &387 &595 &628 \\\cmidrule{1-7}
1 &974 &1980 &559 &154 &403 &384 \\\cmidrule{1-7}
2 &985 &1993 &721 &191 &398 &424 \\\cmidrule{1-7}
4 &992 &1701 &823 &147 &507 &556 \\\cmidrule{1-7}
8 &1006 &1413 &767 &154 &425 &509 \\\cmidrule{1-7}
16 &998 &1435 &724 &189 &427 &451 \\\cmidrule{1-7}
32 &1006 &1232 &692 &241 &487 &491 \\\cmidrule{1-7}
64 &991 &1266 &658 &250 &464 &490 \\\cmidrule{1-7}
128 &1022 &1160 &830 &407 &576 &579 \\\cmidrule{1-7}
256 &1003 &1108 &813 &483 &804 &766 \\\cmidrule{1-7}
512 &1021 &1599 &1534 &1265 &1378 &1421 \\
\bottomrule
\end{tabular}
}
\caption{Number of pair of sentences generated by LLM for LT-Swap across frequency bins using the frequencies of BabyLM100M. Each of these pairs is used to form a quadruplet. WS: WordSwap, IS: InflectionSwap, AS: AgreementSwap.}\label{tab:nb_pairs_swap100M}
\end{table*}

\begin{table*}[!htp]\centering
\resizebox{0.8\textwidth}{!}{
\def\arraystretch{1.1}
\begin{tabular}{crrrrrrr}\toprule
LT-Swap-10M &WS-VERB &WS-NOUN &IS-VERB &IS-NOUN &AG-LONG &AG-SHORT \\\cmidrule{1-7}
0 &NA &NA &3180 &1469 &2258 &2371 \\\cmidrule{1-7}
1 &982 &1988 &937 &305 &810 &910 \\\cmidrule{1-7}
2 &984 &1734 &779 &234 &562 &648 \\\cmidrule{1-7}
4 &1004 &1506 &531 &146 &379 &367 \\\cmidrule{1-7}
8 &1001 &1510 &498 &113 &399 &434 \\\cmidrule{1-7}
16 &1002 &1197 &294 &135 &268 &269 \\\cmidrule{1-7}
32 &1002 &1179 &263 &126 &206 &233 \\\cmidrule{1-7}
64 &997 &663 &242 &155 &233 &250 \\\cmidrule{1-7}
128 &734 &705 &199 &173 &170 &152 \\\cmidrule{1-7}
256 &620 &416 &155 &152 &167 &178 \\\cmidrule{1-7}
512 &742 &1121 &102 &190 &82 &80 \\
\bottomrule
\end{tabular}
}
\caption{Number of pair of sentences generated by LLM for LT-Swap across frequency bins using the frequencies of BabyLM10M. Each of these pairs is used to form a quadruplet. WS:
WordSwap, IS: InflectionSwap, AS: AgreementSwap.}\label{tab:nb_pairs_swap10M}
\end{table*}

\begin{table*}[!htp]\centering
\resizebox{0.9\textwidth}{!}{
\def\arraystretch{1.1}
\begin{tabular}{lrrrrrrrr}\toprule
LT-Swap-10M &WS-NOUN &WS-VERB &IS-NOUN &IS-VERB &AG-LONG &AG-SHORT &average \\\cmidrule{1-8}
gpt-bert &\textbf{0.873} &\textbf{0.874} &\textbf{0.898} &\textbf{0.921} &\textbf{0.838} &\textbf{0.871} &\textbf{0.879} \\\cmidrule{1-8}
MLSM &0.83 &0.814 &0.881 &0.873 &0.747 &0.833 &0.83 \\\cmidrule{1-8}
LSM &0.831 &0.82 &0.875 &0.865 &0.719 &0.825 &0.823 \\\cmidrule{1-8}
babyllama &0.844 &0.83 &0.818 &0.84 &0.672 &0.775 &0.796 \\\cmidrule{1-8}
ltgbert &0.738 &0.734 &0.846 &0.85 &0.62 &0.74 &0.755 \\\cmidrule{1-8}
roberta-base &0.799 &0.79 &0.814 &0.812 &0.55 &0.697 &0.744 \\\cmidrule{1-8}
antlm-bert &0.822 &0.81 &0.792 &0.747 &0.561 &0.656 &0.731 \\\midrule
opt-125M &0.78 &0.784 &0.76 &0.756 &0.52 &0.618 &0.703 \\
\bottomrule
\end{tabular}
}
\caption{LT-Swap scores per subtasks for models pretrained on BabyLM100M. WS: WordSwap, IS: InflectionSwap, AS: AgreementSwap.}\label{tab:scores_per_pos_100M}
\end{table*}

\begin{table*}[!htp]\centering
\resizebox{0.9\textwidth}{!}{
\def\arraystretch{1.1}
\begin{tabular}{lrrrrrrrr}\toprule
LT-Swap-100M &WS-VERB &WS-NOUN &IS-VERB &IS-NOUN &AG-LONG &AG-SHORT &average \\\cmidrule{1-8}
gpt-bert &\textbf{0.907} &\textbf{0.899} &\textbf{0.9} &\textbf{0.916} &0.857 &0.859 &\textbf{0.89} \\\cmidrule{1-8}
ltgbert &0.774 &0.767 &0.88 &0.879 &\textbf{0.869} &0.84 &0.835 \\\cmidrule{1-8}
MLSM &0.753 &0.719 &0.87 &0.869 &0.846 &\textbf{0.886} &0.824 \\\cmidrule{1-8}
opt-125M &0.894 &0.878 &0.819 &0.808 &0.657 &0.767 &0.804 \\\cmidrule{1-8}
babyllama &0.851 &0.828 &0.815 &0.814 &0.763 &0.78 &0.808 \\\cmidrule{1-8}
LSM &0.734 &0.704 &0.851 &0.855 &0.775 &0.869 &0.798 \\\cmidrule{1-8}
roberta-base &0.812 &0.793 &0.829 &0.816 &0.592 &0.687 &0.755 \\\midrule
antlm-bert &0.735 &0.72 &0.742 &0.673 &0.619 &0.688 &0.696 \\
\bottomrule
\end{tabular}
}
\caption{LT-Swap scores per subtasks for models pretrained on BabyLM10M. WS: WordSwap, IS: InflectionSwap, AS: AgreementSwap.}\label{tab:scores_per_pos_10M}
\end{table*}

\begin{table*}
\resizebox{\textwidth}{!}{
\begin{tabular}{l}
\begin{tabularx}{\linewidth}{X}
\hline
\multicolumn{1}{l}{\textbf{WordSwap Noun}} \\
Sentence A: The magician's skillful performance held the audience in a state of complete \textbf{captivation}, mesmerizing them with his incredible illusions and tricks.  \\
Sentence B: The scientists at the research institute used advanced \textbf{spectrography} techniques to analyze the chemical composition of the newly discovered planet. \\
{\textbf{WordSwap Verb}} \\
Sentence A: The renowned actress \textbf{reprized} her iconic role as the lead character in the classic musical, receiving a standing ovation. \\
Sentence B: The highly skilled neurosurgeon carefully \textbf{innervated} the damaged nerve endings to restore full motor function to the patient's paralyzed limb.|42\\
\hline
\multicolumn{1}{l}{\textbf{InflectionSwap Noun}} \\
Sentence A: The recipe calls for one medium-sized \textbf{zucchini} , which should be sliced into thin rounds. \\
Sentence B: The farmers market had a large basket of \textbf{zucchinis} , each one perfectly ripe and ready to eat. \\
{\textbf{InflectionSwap Verb}} \\
Sentence A: The hikers had finally \textbf{surmounted} the steep mountain trail after hours of climbing.\\
Sentence B: Every year , a new group of adventurers \textbf{surmounts} the same challenging peak.\\
\hline
\multicolumn{1}{l}{\textbf{AgreementSwap long-distance subject-verb}}\\
Sentence A: The \textbf{archivist} that can be relied upon to handle fragile documents takes .\\
Sentence B: The \textbf{archivists} that can be relied upon to handle fragile documents take .\\
{\textbf{AgreementSwap short-distance subject-verb}}\\
Sentence A: The \textbf{archivist} takes .\\
Sentence B: The \textbf{archivists} take .\\
{\textbf{AgreementSwap long-distance anaphora}}\\
Sentence A: The \textbf{adventurist} that can be found in the most remote places proved himself .\\
Sentence B: The \textbf{adventurists} that can be found in the most remote places proved themselves .\\
{\textbf{AgreementSwap short-distance anaphora}}\\
Sentence A: The \textbf{adventurist} proved himself .\\
Sentence B: The \textbf{adventurists} proved themselves .\\
{\textbf{AgreementSwap short-distance determinant-noun}}\\
Sentence A: This \textbf{choirboy}.\\
Sentence B: These \textbf{choirboys}.\\
\hline
\end{tabularx}
\end{tabular}
}
\caption{Example generations for LT-Swap100M.}
 \label{table:sentence_examples}
\end{table*}

\begin{table*}
\resizebox{\textwidth}{!}{
\def\arraystretch{1.1}
\begin{tabular}{l}
\begin{tabularx}{\linewidth}{X}
\hline
{\textbf{Nouns}}\\
$1$ : landsman garrulity cusk pitchman relator sternum greensward oryx mineralogy virtuoso \\
$[2,4[$ : fallow covey plurality codeine bushman poesy quince chancery quinsy backgammon \\
$[4,8[$ : bluebird kilt fourpence yap tapestry pinny butchery conformance chaise sexist \\
$[8,16[$ : buckwheat reeve buckwheat bloc buckwheat forester carp droning fanfare spaceflight \\
$[16,32[$ : merchandise ape meteor churchyard severity hub dew lineage grocer incline \\
$[32,64[$ : revival activism shotgun baptism squire diesel clover bandaid creepy terrazzo \\
$[64,128[$ : bride breach porridge stake combat imagination costume ichinomiya contrast devotion \\
$[128,256[$ : intake stove chamber arrondissement height donkey discussion chin description hammer \\
$[256,512[$ : prince theory difficulty campaign mayor cash ocean mail valley wolf \\
$[512+[$ : number cup television attack sight duck chance company country truck \\
\hline
{\textbf{Verbs}}\\
$1$ : codify yammer lengthwise introspect heliolatry phrensy betaken bilk overreach clayey \\
$[2,4[$ : redid therefor snuffy therefor meathead firsthand almshouses certes comport misspoke \\
$[4,8[$ : wiggled outlived dictated accentuated embossed battled certificated disadvantaged resuscitated amputated \\
$[8,16[$ : exchanging converging videotaping alluring resolving videotaping resolving alluring rearranging videotaping \\
$[16,32[$ : securing clasping typing prowling decorating inquiring interfering completing cuddling clasping \\
$[32,64[$ : ignoring scoring scratching supposing scoring challenging ignoring challenging scoring investigating \\
$[64,128[$ : acquired commenced struggled complicated graduated ejaculated rescued counted complicated commenced \\
$[128,256[$ : associated inquired seized ceased claimed supported compared connected divided shed \\
$[256,512[$ : distributed realized split hidden sang hidden split sang facing kidding \\
$[512+[$ : lay feel came put stay drew wear grew knew stop \\
\hline
\end{tabularx}
\end{tabular}
}
\caption{Example nouns and verbs per frequency bins found in BabyLM 100M.}
 \label{table:word_examples}
\end{table*}

\clearpage
\newpage

\section{Tokenization analysis}\label{appendix:tokenization}

\subsection{Context}

Although subword units—such as BPEs and WordPiece—are the most widely used form of tokenization in language models \cite{brown2020gpt3,geminiteam2024gemini,grattafiori2024llama3}, some recent architectures have begun to eliminate tokenizers altogether, operating directly on character sequences \cite{canine2021,yu2023megabytes,pagnoni2024blt}. Character-level LMs present a double-edged sword. On one hand, they are expected to generalize better than subword-based models on certain tasks, as they have direct access to word spellings. On the other hand, character-level tokenization significantly increases sequence length, which may impair performances on semantic tasks requiring efficient learning of word co-occurrences.

This trade-off motivates our hypothesis in Section \ref{sec:lt-swap-correlations}: language models that explicitly separate syntactic markers from their base words should perform better on syntactic tasks (InflectionSwap and AgreementSwap), \textit{particularly} when dealing with rare words. However, WordSwap performance may suffer due to the increased sequence length introduced by character-level tokenization.

To evaluate this hypothesis, we train language models on the BabyLM datasets using a range of tokenization strategies, spanning characters, subwords (BPEs), and full words. 

%\cite{mikolov2011SUBWORDLM,graves2013charlevel,bojanowski2015alternativestructurescharacterlevelrnns,nguyen2022wordboundaries}. 

%While character-level tokenization offers certain benefits, including robust handling of out-of-vocabulary words and greater morphological flexibility, it introduces substantial challenges in practice. Additionally, learning meaningful linguistic representations from raw characters alone requires the model to capture word dependancies from a more granular. In contrast, words and subword-level methods such as Byte-Pair Encoding (BPE) facilitate more efficient learning. 

\begin{table}[!htp]\centering
\scriptsize
\begin{tabular}{lccccc}\toprule
&\multicolumn{2}{c}{BLiMP} &&\multicolumn{2}{c}{LT-Swap}  \\\cmidrule{2-3}\cmidrule{5-6}
&10M &100M &&10M &100M  \\\cmidrule{2-6} 
words &0.611 & 0.738 &&0.720& 0.796\\
BPE-50k* &0.602 & 0.733 &&0.748&0.822\\
BPE-50k &0.582 & 0.752 &&0.736&0.835\\
BPE-16k &0.643 & 0.741 &&0.767&0.843\\
chars &0.646 & 0.721 &&0.818	&0.857\\
\bottomrule
\end{tabular}
\caption{BLiMP and LT-Swap scores (i.e average of WordSwap, InflectionSwap and AgreementSwap) for LM trained with different tokenization methods (words, BPE, characters) on either BabyLM10M and BabyLM100M. BPE-50k* is the score obtained by the baseline provided by the BabyLM organizers while BPE-50k is our attempt at replicating this baseline.}\label{tab:blimp_tokens}
\end{table}

\subsection{Experiments}

Here are the details of our training setups for character, BPE and word tokenizers. Character-level tokenization is obtained by splitting text into individual characters. For BPEs, we trained BPE models on the BabyLM datasets using 16k units as done by the GPT-BERT authors \cite{charpentier2024gptbert}. In addition, in order to replicate the baseline OPT-125M provided by the BabyLM organizers, we took a 50k units BPE tokenizer pretrained on WebText by \citet{radford2019LanguageMA}. Finally, in order to train a word-level language models with a manageable vocabulary size, we lower-cased the BabyLM datasets, we put white space around all symbols and figures and we put to "<UNK>"  all words longer than 5 letters that did not belong to the English dictionnary. By doing so we reach a vocabulary size of 69384 for BabyLM10M and 131059 for BabyLM100M. After checking, the number of word tokens that are in the BabyLM datasets but not in the BLiMP or LT-Swap is marginal (even in the rarest frequency bins of LT-Swap).

For each tokenization method and each BabyLM dataset (10M and 100M), we trained an OPT-125M transformer encoder \cite{zhang2022opt} with 12 layers and hidden size 768, using Adam optimizer, 250 warm up steps, peak learning rate 0.001 and cosine decay down to 0.0001. The batch size is set to 0.5M tokens and sequence length to 128 for words and BPEs and 512 for characters. We stopped training when the loss on the BabyLM dev set stopped decreasing.

\subsection{Results}

\subsection{Overall scores}

\begin{figure*}[h!]
        \centering
        \includegraphics[width=.45\textwidth]{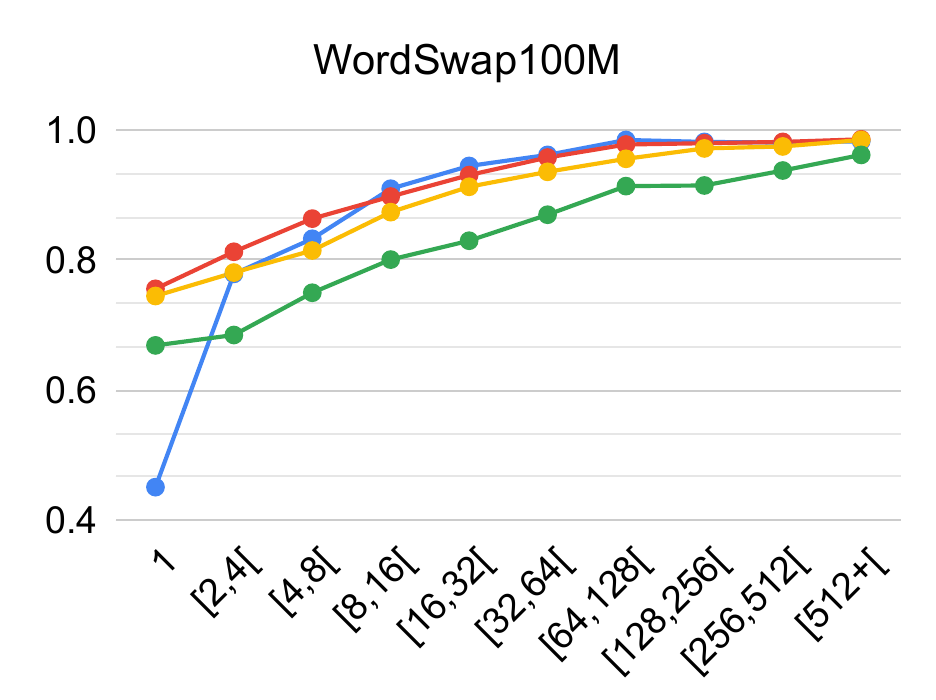}\hfill
        \includegraphics[width=.55\textwidth]{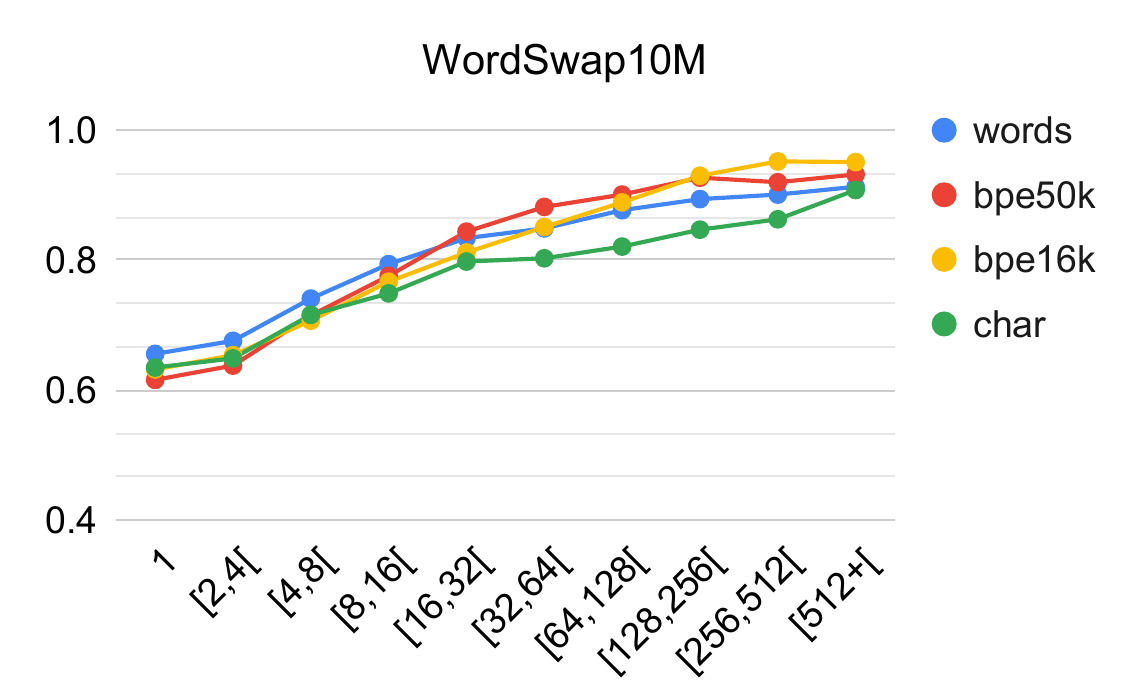}
        \includegraphics[width=.45\textwidth]{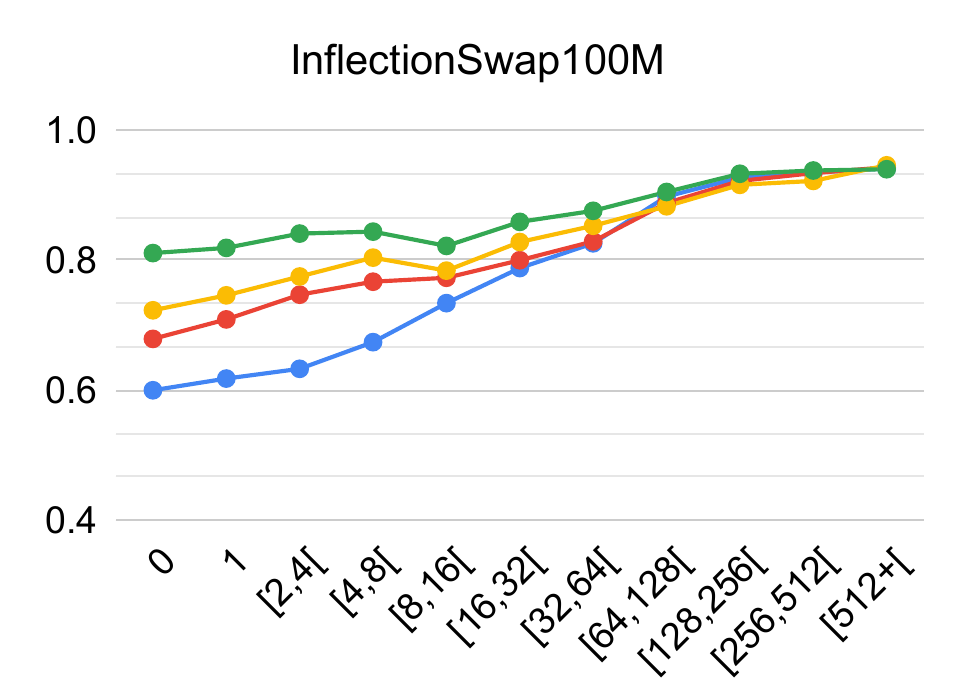}\hfill
        \includegraphics[width=.55\textwidth]{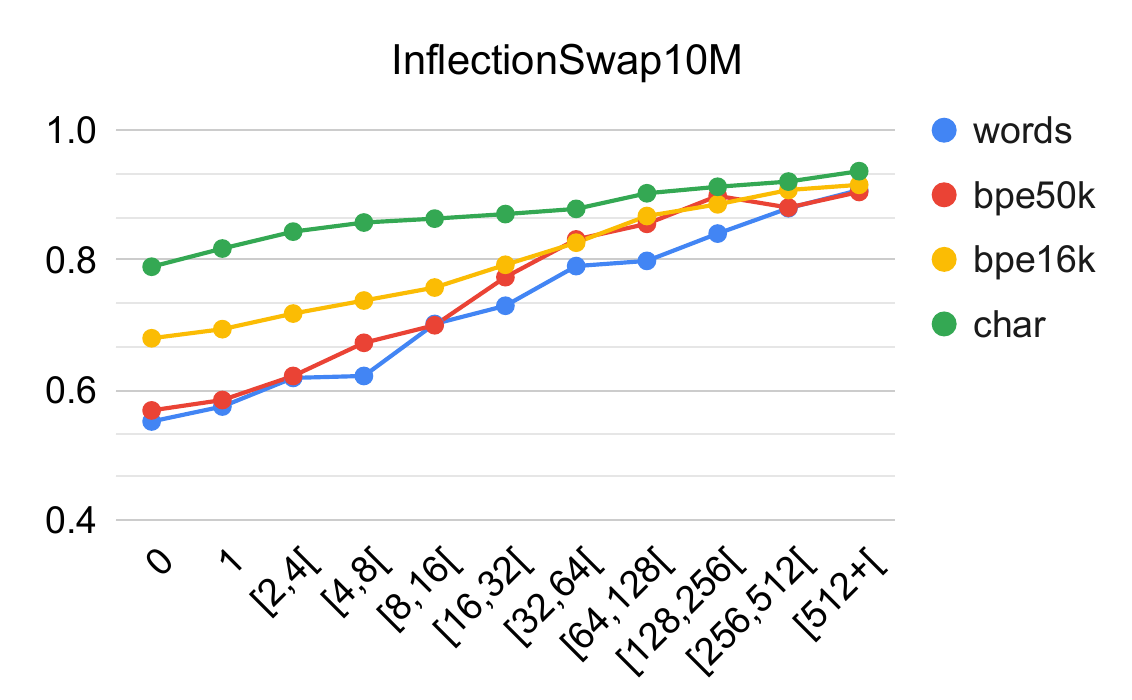}
        \includegraphics[width=.45\textwidth]{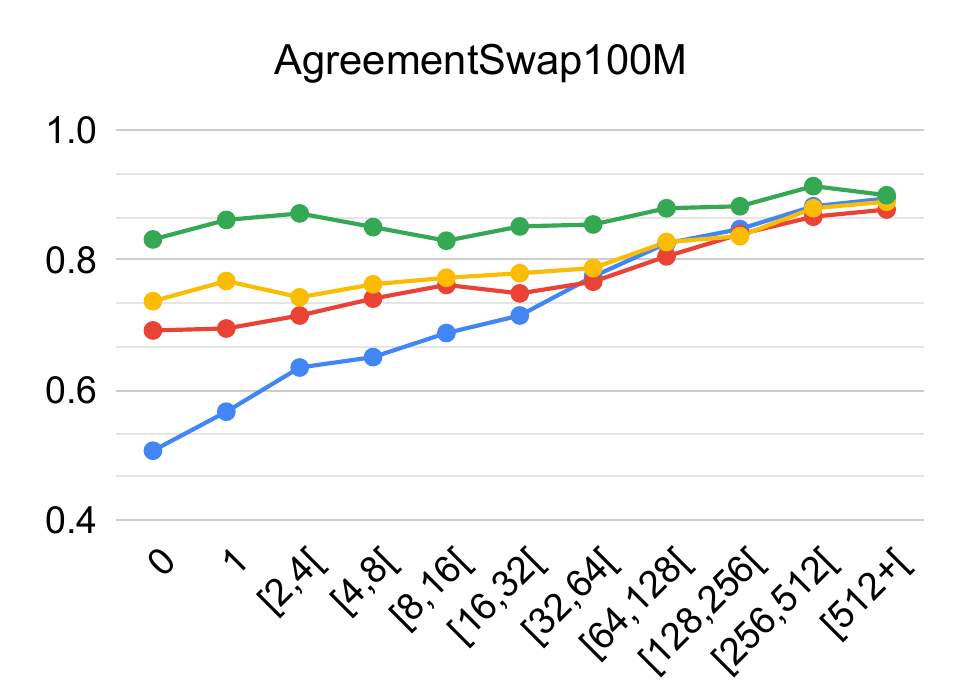}\hfill
        \includegraphics[width=.55\textwidth]{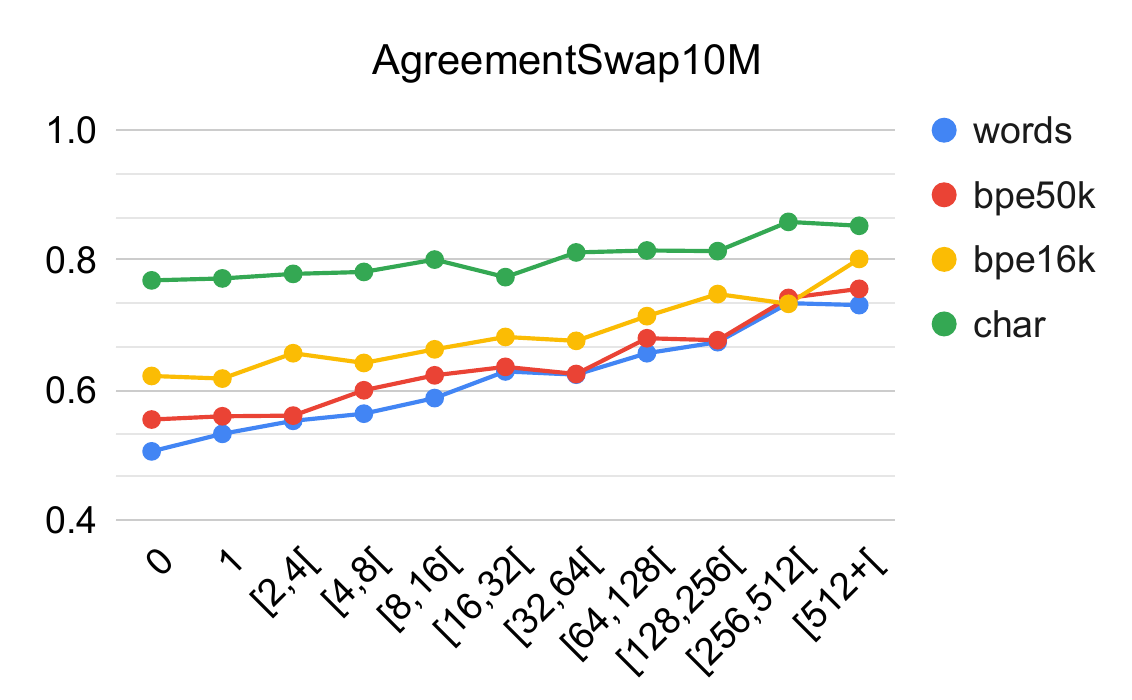}
        \caption{Scores obtained on our three LT-Swap tasks over BabyLM10M and BabyLM100M for four different tokenization methods: char-level, bpe (16k and 50k units) and words (i.e white space segementation). The standard error for all Swap scores is less than 0.02.}
\label{fig:score_per_tokenisation}
\end{figure*}

LMs performances on BLiMP and LT-Swap are presented in Table \ref{tab:blimp_tokens}. From those results, we first notice that our training procedure replicates the OPT-125M baseline provided by the BabyLM organizers with scores. Second, while BLiMP scores do not show a clear trend across tokenization method, the LT-Swap benchmark reveals a clear correlation: the lower the vocabulary size, the better the scores. In order to explain this trend, we present the detailed LT-Swap scores across frequency bins and subtasks in Figure \ref{fig:score_per_tokenisation}. 

The first observation is that once again LMs performances are more evenly distributed for rare words than for frequent words. This explains why BLiMP, which focuses on the head of the word distribution instead of the tail, do not show clear differences across tokenization methods. 

Second, for the semantic task (WordSwap) high vocabulary size generaly gives better performances (except for the hapax case which we detail in the next section). Yet, the opposite is happening very clearly on syntactic tasks (measured by InflectionSwap and AgreementSwap) where the lower the vocabulary size the higher the performance.  

These results confirm our hypothesis from Section~\ref{sec:lt-swap-correlations}. Subword and word-level tokenizations outperform character-level models on semantic tasks, likely because larger token units make it easier to learn word co-occurrence patterns—world knowledge and concepts can often be captured by a single or a small number of embeddings. However, character-level language models consistently outperform subword and word-level models on syntactic tasks.

More notably, character-level LMs exhibit only a mild performance drop between high- and low-frequency words on syntactic tasks, while showing a substantial drop on the semantic WordSwap task. We hypothesize that this robustness stems from the character-level model’s ability to isolate syntactic cues (e.g., -ed for past tense, -s for plural) from the base word. As a result, the model can effectively perform syntactic transformations even when the base word appears rarely—or not at all—during training.

\subsection{Hapaxes and word-level tokens}

Another unexpected result is the sharp performance drop (down to random chance) of the word-level language model on the WordSwap-100M dataset specifically for the hapax case—words that appear only once. This degradation is notably absent on the smaller WordSwap-10M dataset. This suggests that in small enough corpora, even a single occurrence of a word can suffice to derive a useful semantic embedding. However, as the dataset size grows while the model size remains fixed, it becomes increasingly difficult for the model to retain information about such rare words. In effect, the model may "forget" what it has learned about hapaxes due to capacity constraints.

\end{document}